\documentclass[11pt]{article}

\usepackage[final]{acl}

\usepackage{latexsym}
\usepackage{booktabs}
\usepackage{amsmath}
\usepackage{array}

\usepackage{fontspec}
\usepackage{xeCJK}
\usepackage{polyglossia}

\usepackage{alltt}
\setmainlanguage{english}
\xeCJKsetup{CJKspace=true}
\setotherlanguage{arabic}
\newfontfamily\arabicfont{Amiri-Regular.ttf}[
  BoldFont       = Amiri-Bold.ttf,
  ItalicFont     = Amiri-Italic.ttf,
  BoldItalicFont = Amiri-BoldItalic.ttf,
  Script         = Arabic ]

\usepackage{microtype}

\usepackage{placeins}

\usepackage{graphicx}

\title{Personas Differ from Native-Language Generation: Language Pathways Shape LLM Interpersonal Advice}

\author{Jinhee Won \\
  Massachusetts Institute of Technology \\
  \texttt{jinhee@mit.edu} \\\And
  Xinlan Emily Hu \\
  Massachusetts Institute of Technology \\
  \texttt{xehu@mit.edu} \\}

\begin{document}
\maketitle
\begin{abstract}
LLMs are increasingly used for interpersonal advice and as tools for studying social behavior across languages and cultures. A common shortcut for eliciting language- or culture-related variation is to ask a model to answer as a native speaker. We test whether this native-speaker persona reproduces the outputs obtained when models instead generate advice in the target language and translate the response back into English. Using 600 interpersonal advice questions across 13 languages and eight LLMs, we compare native-language generation followed by translation (NL) with native-speaker persona prompting (NP), measuring linguistic style, behavioral scaffolding, and forced-choice action recommendations. We find that NP and NL are not interchangeable. Compared to NL, NP often increases lexical social cues, including affiliation and positive tone, while reducing qualities such as concreteness and social attunement; NP also provides less actionable scaffolding in open-ended advice. In forced-choice scenarios, NP changes which action the model selects, favoring confrontation over redirection, with effect sizes varying across languages, topics, and models. Our results show that cross-lingual elicitation strategy is a consequential methodological choice that can change both how advice is framed and which actions models recommend.
\end{abstract}

\section{Introduction}

\begin{figure}
    \centering
    \includegraphics[width=1\linewidth]{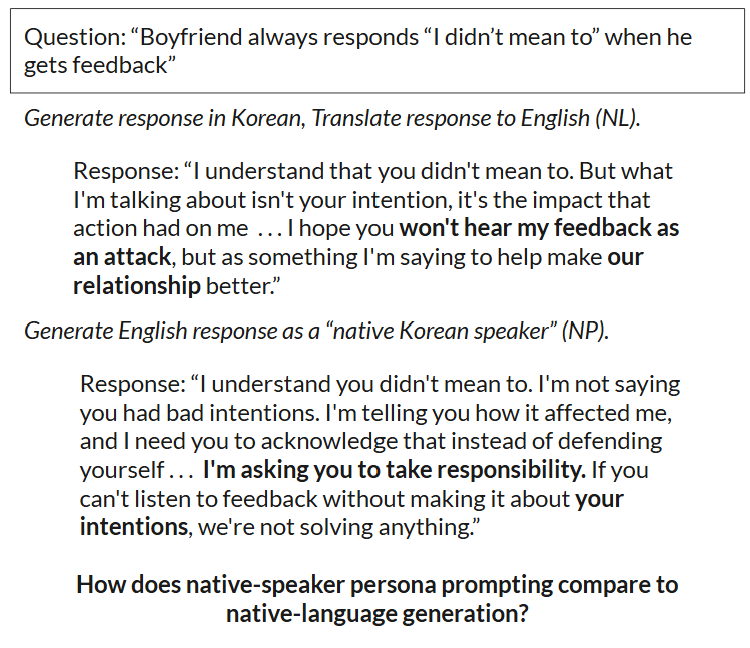}
    \caption{Example of how GPT-5.5 varies based on prompting strategy.}    
    \label{fig:motivating_example}
\end{figure}

Large language models (LLMs) are widely used for interpersonal guidance, from workplace communication and boundary-setting to relationship conflict and family disagreements~\citep{chatterji2025how}. In these settings, models do more than produce language: they help users decide how to behave toward others. This is especially consequential in cross-lingual and culturally mismatched contexts, where users may know \textit{what} they want to say but not \textit{how} to say it. Because languages and cultures differ in norms of directness, formality, emotion, deference, and values, interpersonal advice tightly links linguistic form with social meaning.

\begin{figure*}[t]
\centering
\includegraphics[width=\textwidth]{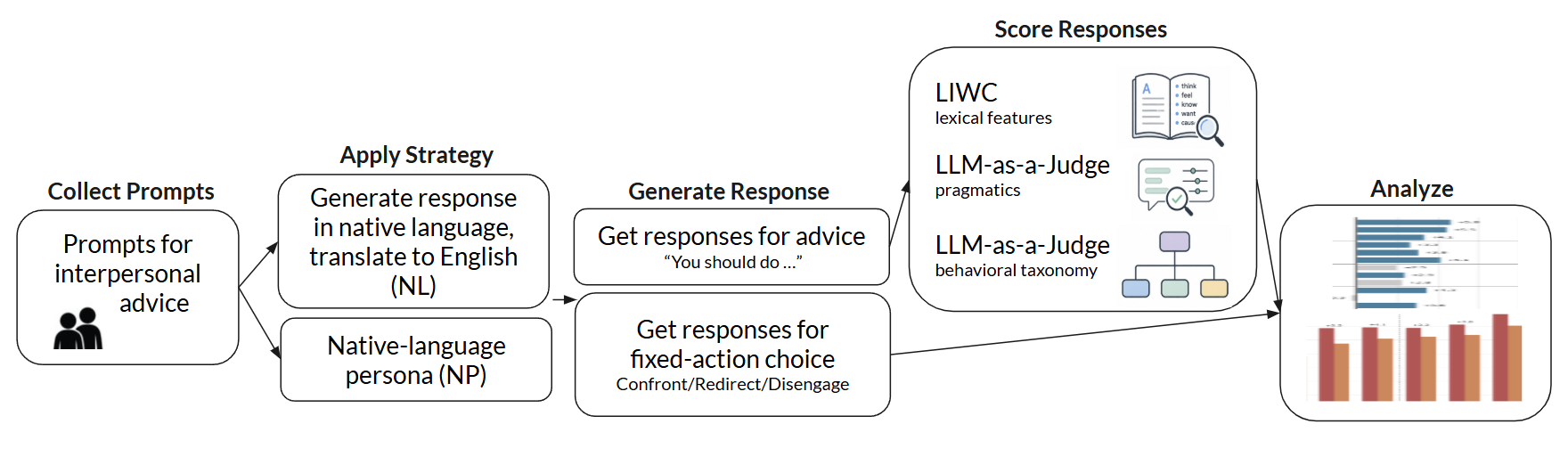}
\caption{Framework for evaluating how language pathways affect LLM-generated responses to interpersonal advice prompts.}
\label{fig:methodology}
\end{figure*}

However, multilingual LLM evaluation has focused largely on objective NLP benchmarks such as translation, question answering, reasoning, and instruction following~\citep{hu2020xtreme,ahuja2023mega,lai2023chatgpt,singh2024aya,ustun2024aya}. These tasks do not fully capture interpersonal advice, where there may be no single correct answer and differences in directness, politeness, hierarchy, and emotional expression can change the force of a response~\citep{goffman1967interaction,brown1987politeness,tingtoomey1998facework}. This limitation is especially important because LLMs are increasingly used to model human behavior through demographic backstories, persona variables, prompt language, and cultural frames~\citep{argyle2023out,hu2024quantifying,tao2024cultural}, even though recent work shows that persona prompts can influence outputs without reliably capturing the full complexity of human subjectivity or cultural variation~\citep{hu2024quantifying,giorgi2024modeling}. In multilingual interpersonal advice, this raises a methodological question: is asking a model to respond as a "native speaker" equivalent to asking it to process or generate advice in the target language?

This study investigates how language pathways shape LLM-generated interpersonal advice (Figure~\ref{fig:motivating_example}). Starting from English interpersonal advice questions, we translate prompts into twelve languages and compare responses from five open-weight and three proprietary API models. We compare native-language generation followed by translation into English (NL) with native-speaker persona prompting in English (NP) to test whether these two elicitation pathways produce equivalent advice.

Our analysis examines both how advice is framed and what the model recommends. We measure linguistic characteristics, including tone, power, formality, and concreteness; behavioral guidance using the Behavior Change Technique Taxonomy~\citep{michie2013behavior}; and recommended action through a forced-choice classification of confrontation, disengagement, or redirection.

This paper makes three contributions. First, we show that language pathway is not a neutral implementation detail: the same interpersonal dilemma yields different linguistic and behavioral outputs depending on whether the model generates in the target language or is prompted with a native-speaker persona in English. Second, we show that NP and NL diverge along both linguistic and behavioral dimensions. NP often adds lexical social cues, such as prosocial language and positive tone, while reducing concreteness, social attunement, emotional expressiveness, and structured behavioral guidance relative to NL. Third, we show that these differences vary by language, topic, and model, suggesting structured elicitation effects rather than random variation. These patterns should be viewed as properties of model elicitation; they are not evidence about real language communities. Studying the latter would require validating which of NP or NL better aligns with responses from actual speakers of the language.

\section{Related Work}

\subsection{Multilingual LLM Evaluation}

Multilingual NLP evaluation has traditionally focused on whether models transfer across languages on standard tasks such as classification, question answering, retrieval, translation, and structured prediction. Benchmarks such as XTREME evaluate multilingual representations across many languages and task types \citep{hu2020xtreme}. More recent LLM-oriented benchmarks such as MEGA and ChatGPT Beyond English evaluate generative models across multilingual settings \citep{ahuja2023mega,lai2023chatgpt}. Multilingual instruction-tuning efforts such as Aya address the English-centered nature of LLM development by constructing multilingual instruction datasets and models \citep{singh2024aya,ustun2024aya}. While existing literature shows that LLM behavior varies substantially across languages, most evaluations emphasize task correctness, benchmark performance, or instruction-following quality. Interpersonal advice differs because there may be no single correct answer; the relevant question concerns interpersonal stance, tone, and action recommendation.

\subsection{LLMs for Interpersonal Advice and Social Judgment}

A growing body of work examines LLMs in social, interpersonal, and emotional-support settings. Studies of LLM-generated advice show that advice quality is shaped not only by the recommendation itself, but also by phrasing, tone, and perceived usefulness~\citep{wester2024exploring}. Work on social-situational judgment finds that LLMs can perform strongly on scenarios involving interpersonal conflict and social reasoning~\citep{mittelstadt2024social}. Related work on emotional support studies whether LLMs can produce empathic or culturally sensitive responses, while showing that simple role-play prompts may be insufficient for cultural sensitivity~\citep{liu2026tailored}.

\subsection{Personas, Cultural Prompting, and Social Bias}

Increasingly, LLMs are used as proxies for human respondents or social actors by conditioning them on demographic, cultural, or persona descriptions. Persona prompts affect model outputs but capture only part of human variation and context~\citep{hu2024quantifying,giorgi2024modeling}. Persona choices can shift cultural-norm judgments~\citep{kamruzzaman2026woman}, while broader bias research documents demographic stereotypes across languages and variation in judgments and persuasive style by gender and relationship attributes~\citep{nadeem2021stereoset,smith2022holisticbias,costajussa2023multilingual,kotek2023gender,levy2024gender,pauli2026analysing}.

Prompt language and cultural framing provide additional ways of conditioning whose effects depend on how culture is represented and evaluated~\citep{alkhamissi2024investigating}. Generating and combining culturally prompted responses across languages can increase demographic and perspective diversity~\citep{wang2025multilingualpromptingimprovingllm}, while structured value surveys find only limited gains in alignment with human cultural values~\citep{bulte2025llmsculturalvaluesimpact}.

\section{Methodology}

\subsection{Data Collection}

We collect interpersonal advice questions from the Interpersonal Skills Stack Exchange, a public forum focused on everyday interpersonal problems~\citep{stackexchange_interpersonal}. We select the top 150 questions from four tags: \texttt{work-environment}, \texttt{friends}, \texttt{relationships}, and \texttt{family}, yielding 600 English questions. Sample prompts are shown in Appendix~\ref{app:advice_question_samples}. These questions are advice-seeking, socially situated, and behaviorally open-ended, making them suitable for evaluating differences in tone, directness, politeness, social orientation, and recommended action.

\subsection{Languages}

We evaluate 13 languages spanning multiple scripts, language families, and regions: Japanese, Korean, Chinese, Arabic, Persian, Turkish, English, German, Dutch, Swedish, Spanish, Italian, and Portuguese. English uses the original questions, while all other languages use the translation pipeline below to construct semantically controlled non-English versions. As descriptive aids for our exploratory analysis, we group these languages into broad categories such as East Asian, Middle Eastern, Germanic, and Romance.

\subsection{Question Translation}

We translate the English questions into the 12 non-English languages using \texttt{facebook/nllb-200-distilled-1.3B}, a multilingual sequence-to-sequence model from No Language Left Behind \citep{costajussa2022nllb,meta2022nllb200distilled13b}. For each question-language pair, we generate 32 candidate translations with temperature 0.3 and back-translate each candidate into English four times with temperature 0.7. We then select the candidate with the minimum variance among the five highest-mean candidates in cosine similarity to the original question, computed using \texttt{all-MiniLM-L6-v2} \citep{reimers2019sentencebert, sentenceTransformers2021allminilml6v2}. Selected translations show high semantic preservation overall, with mean similarity of 0.928 across languages and language-level means ranging from 0.876 for Japanese to 0.976 for Spanish; additional diagnostics and examples appear in Appendix~\ref{app:translation_pipeline}.

\subsection{Response Generation}
We generate responses from five open- and three proprietary models:
\texttt{google/gemma-4-E4B-it}, \texttt{google/gemma-4-26B-A4B-it},
\texttt{google/gemma-4-31B-it}, \texttt{Qwen/Qwen3.6-27B},
\texttt{Qwen/Qwen3.6-35B-A3B}, \texttt{claude-opus-4.6},
\texttt{gpt-4o}, and \texttt{gpt-5.5} \citep{google2026gemma4, qwen2026qwen36_27b, qwen2026qwen36_35b_a3b,
anthropic2026opus46, openai2024gpt4o, openai2026gpt55}. For each non-English language, model, and strategy, we generate responses to all 600 questions using four prompting strategies (our two primary strategies of NL and NP, plus two auxiliary strategies discussed in Appendix~\ref{app:generation_prompts}). This yields 2,400 records per model per non-English language. For English, models answer the original English questions.

For each non-English language, our primary comparison is between \textit{native language generation} (NL), where the model receives and answers the question in the target language before the answer is translated into English for analysis, and \textit{native-speaker persona prompting} (NP), where the model receives the original English question and answers in English as a  "native speaker" of the target language. Exact prompts are provided in Appendix~\ref{app:generation_prompts}.

\subsection{Annotation and Measurement}

We analyze generated responses using four measurement layers: LIWC-22 lexical features, LLM-based pragmatic annotations, LLM-based behavior-change technique classifications, and forced-choice behavioral coding.

First, we compute 12 LIWC lexical features on the English-analyzable final responses. LIWC provides dictionary-based measures of psychologically and socially meaningful word use~\citep{tausczik2010psychological,pennebaker2015development,boyd2022development}. We use these measures to capture social, affective, cognitive, and motivational language: \textit{affiliation}, \textit{achievement}, \textit{power}, \textit{cognitive processes}, \textit{certitude}, \textit{tentativeness}, \textit{emotion}, \textit{positive tone}, \textit{negative tone}, \textit{social references}, \textit{prosocial language}, and \textit{conflict}.

Second, we use \texttt{gpt-4o} to annotate holistic properties of the advice along five dimensions that are difficult to capture with lexical counts alone. \textit{Directness} captures how explicitly recommendations are communicated; \textit{Formality} captures how advice signals social distance and situational expectations \citep{biber1995dimensions}. \textit{Emotional expressiveness} captures how overtly advice conveys affect \citep{dubois2007stance}; \textit{Social attunement} captures attention to the recipient's feelings and relational concerns \citep{goffman1967interaction,brown1987politeness}; and \textit{Concreteness} captures the distinction between abstract guidance and specific details \citep{trope2010construal}. The evaluator scores each dimension on a five-point ordinal scale. A score of 1 indicates indirect, informal, emotionally restrained, task-focused, or abstract language, respectively, whereas a score of 5 indicates direct, formal, emotionally expressive, socially attuned, or concrete language. Intermediate scores represent graded positions between these endpoints. We use LLM-based annotation because these properties are context-sensitive; prior work shows that structured LLM evaluators can support scalable assessment of open-ended text, while requiring caution about evaluator bias \citep{liu2023geval,zheng2023judging}.

Third, we code responses using higher-level clusters adapted from the Behavior Change Technique Taxonomy \citep{michie2013behavior} with an LLM-as-a-Judge. Although the taxonomy was developed for behavior-change interventions, several categories capture general forms of action guidance that also appear in interpersonal advice: \textit{identifying a problem}, \textit{planning a concrete response}, \textit{anticipating consequences}, \textit{reframing the situation}, \textit{seeking support}, and \textit{monitoring outcomes}. The BCT categories therefore provide a useful structure for measuring behavioral scaffolding in advice, which we define as guidance that helps the recipient select, plan, carry out, and evaluate a course of action. The evaluator scores each BCT category on a five-point ordinal scale, where 1 indicates that the technique is absent and 5 indicates that it is strongly present; intermediate scores indicate increasing levels of support within that category. This complements linguistic analysis, as two responses may sound equally polite or prosocial while differing in whether they help the user decide what to do, how to do it, and what consequences to consider.

Finally, we run a separate forced-choice action task motivated by frameworks that distinguish direct engagement, avoidance, compromise, and integrative responses~\citep{rahim1983measure,gross2000managing}. For each question-language-strategy condition, the model receives the same interpersonal dilemma and chooses one of three broad action families: \textit{confrontation}, \textit{disengagement}, or \textit{redirection}. Confrontation directly addresses the issue with the other person or an authority figure; disengagement pauses, reduces, or avoids interaction; and redirection captures non-avoidant but less directly confrontational alternatives such as reframing, changing communication channels, seeking mediation, or pursuing compromise. This direct action choice is distinct from the model's open-ended response.

\subsection{Statistical Analysis}

Our main analysis compares NL and NP across output measures, using English as a reference condition for scaling and visualization. For continuous LIWC, LLM-annotated, and BCT features, we report standardized treatment-minus-English differences for the same source prompt and model. NP--NL tests use matched prompt--model observations: within each source prompt and model, we average feature scores across target languages for each strategy, compute the NP-minus-NL difference, and test whether the mean differs from zero using a one-sample paired-difference \(t\)-test~\citep{student1908probable}. For exploratory language-group analyses, we compute the same NP--NL contrasts within matched observations, average them within broad language groups, and test whether the persona--generation gap differs across groups. For forced-choice action outcomes, each response is labeled as confrontation, redirection, or disengagement. We report strategy-level action distributions as percentages and test whether prompting strategy changes the overall distribution using Pearson chi-square tests over the three action categories~\citep{pearson1900criterion}. 

Across analyses, stars indicate significant differences ((q<0.05)) for the comparison specified in each table or figure. We control the false discovery rate within each comparison family using the Benjamini--Hochberg procedure~\citep{benjamini1995controlling}. Continuous effects are reported in standardized units using English as the reference for scaling; action rates are reported as percentages, and differences in action rates as percentage points.

\section{Linguistic Analysis}

This section analyzes how the cross-lingual prompting strategy affects the linguistic features in the resulting advice using LIWC lexical features and LLM-based annotations. The motivating example in Figure~\ref{fig:motivating_example} illustrates the difference between NP and NL linguistically. The Korean-generated response translated into English (NL) is more relationally attuned, softening with phrases like "I hope you won't hear my feedback as an attack." In contrast, the native-speaker persona (NP) response uses a firmer accountability frame, warning the boyfriend that "if you can't listen to feedback without making it about your intentions, we're not solving anything." This is reflected in the LLM-annotated holistic properties: NP is less formal (2 vs.\ 3) and less socially attuned (3 vs.\ 5).

\subsection{Overall Linguistic Patterns}
\begin{figure}
    \centering
    \includegraphics[width=1\linewidth]{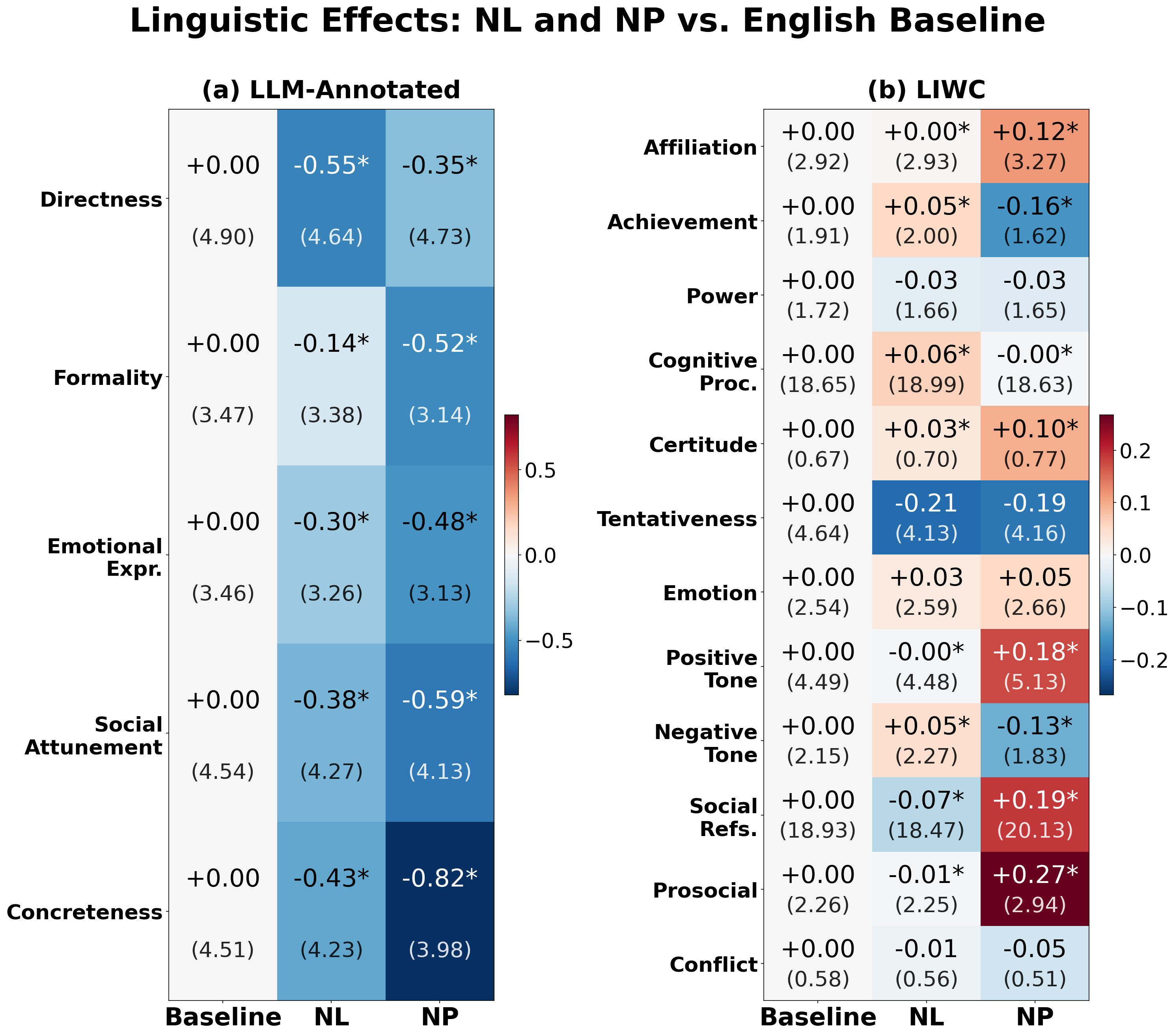}
    \caption{Linguistic features for NP and NL (left: LLM-annotated, right: LIWC), aggregated across responses from 600 questions, 12 non-English languages, and eight models. Main values are normalized relative to the English baseline; values in parentheses report mean scores on the original LLM-annotation and LIWC scales. Stars mark differences significant at $q<0.05$. Compared to NL, NP increases surface social cues such as affiliation and reduces action-oriented features such as achievement and concreteness.}    
    \label{fig:overall_linguistic}
\end{figure}

\begin{figure*}
    \centering
    \includegraphics[width=1\linewidth]{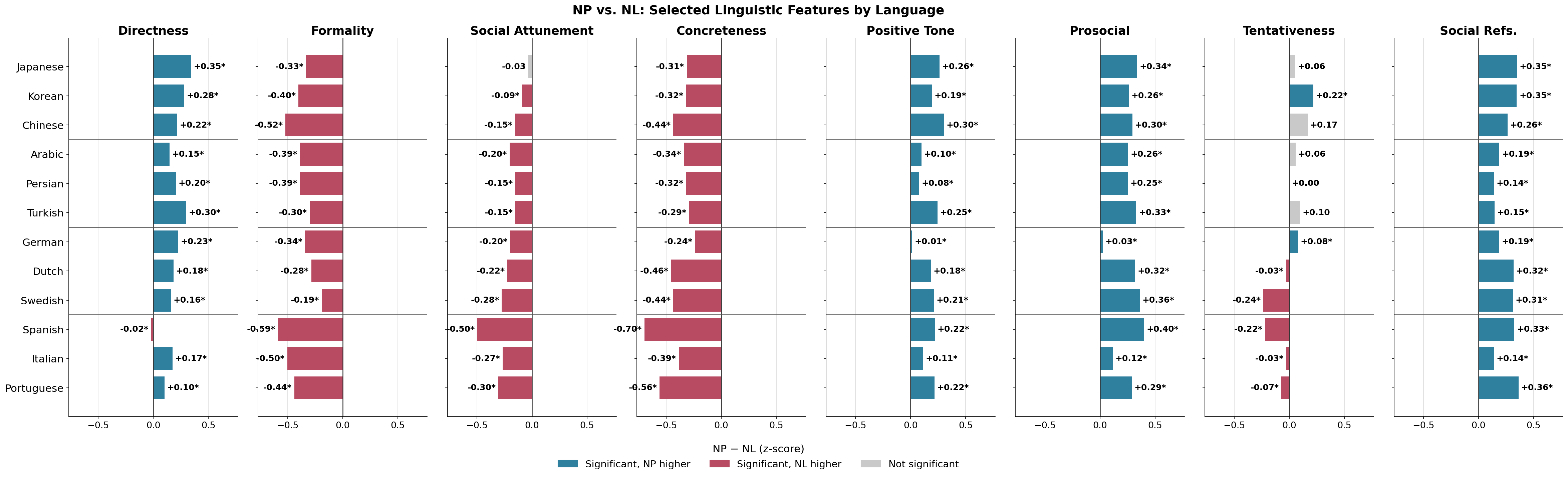}
    \caption{
    NP--NL differences in selected linguistic features by language, aggregated across responses from 600 questions and eight models. The selected features show a subset of how the advice positions the user interpersonally. Differences vary across target languages, showing that persona prompting does not affect all linguistic contexts uniformly.
    }   
    \label{fig:linguistic_lang}
\end{figure*}

Figure~\ref{fig:overall_linguistic} shows that NL and NP produce significantly different linguistic profiles across the 12 foreign languages we study. Overall, NP shifts advice toward an assertive-supportive style: it uses more affiliation, social references, prosocial language, and positive tone, but is less formal, less emotionally expressive, less socially attuned, and less concrete than NL. By comparison, NL produces advice with more contextual and relational elaboration, including greater attention to emotional nuance, relationship dynamics, and concrete interpersonal detail. This overall pattern is broadly consistent across LLMs, with model-level results reported in Appendix~\ref{app:cross_model}.

This contrast highlights a surprising divergence between lexical social cues and pragmatic content. NP uses more language associated with support and affiliation while offering less socially attuned and concrete advice. This indicates that these additional social cues do not translate into greater attention to the recipient’s feelings, relational concerns, or specific circumstances, highlighting another difference between NP and NL.

\subsection{Strategy Differences Across Languages}

The size and direction of NP--NL differences vary systematically by language (Figure~\ref{fig:linguistic_lang}). Across languages, NP generally increases directness and socially positive cues, while NL is higher on formality, social attunement, emotional expressiveness, and concreteness. However, the magnitude of these gaps is not uniform: the NP–NL difference in social attunement and concreteness is smaller for Japanese, Korean, and Chinese than for several Romance and Germanic languages. Exploratory language-group analyses confirm significant differences in effect sizes across broad language groups; full results appear in Appendix~\ref{app:linguistic_lang_group}.

These uneven gaps matter for cross-language comparisons because substituting NP for NL does not affect every language equally. In downstream analyses, this substitution can widen or narrow observed differences between languages and change conclusions about how model behavior varies across them. 

\subsection{Discussion}

Together, the linguistic results show that NP and NL are distinct elicitation methods for multilingual interpersonal advice. NP tends to produce advice that is warmer, more positive, and more explicitly socially marked, while NL more often preserves the relational and pragmatic scaffolding of advice: how directly the user should speak, how much emotional context is included, how carefully the relationship is managed, and how concretely next steps are specified. The difference is not just a stylistic one. Language pathway changes how the advice positions the user in the interpersonal situation.

\section{Behavioral Analysis}

This section examines whether language pathway changes not only how advice sounds, but what it helps the user do. We first analyze generated responses with the Behavior Change Technique Taxonomy (BCT), which captures behavior-guiding components such as problem-solving, planning, reframing, and consequence reasoning. We then use a forced-choice action task to test whether the model ultimately recommends confrontation, disengagement, or redirection. The motivating example in Figure~\ref{fig:motivating_example} illustrates the distinction: NL provides more behavioral scaffolding than NP---that is, more guidance for putting advice into practice. The NL response scores higher on cognitive reframing (4 vs.\ 2), social support (4 vs.\ 2), and problem solving (5 vs.\ 3). For example, NL frames feedback as "I'm saying to help make our relationship better," while NP says the boyfriend needs to "acknowledge that instead of defending." The forced-choice label also reflects the differences between NP and NL, with the Korean generation selecting "Redirection" and the Korean-speaker persona selecting "Confrontation."

\subsection{Behavioral Guidance in Generated Advice}

\begin{figure}
    \centering
    \includegraphics[width=0.75\linewidth]{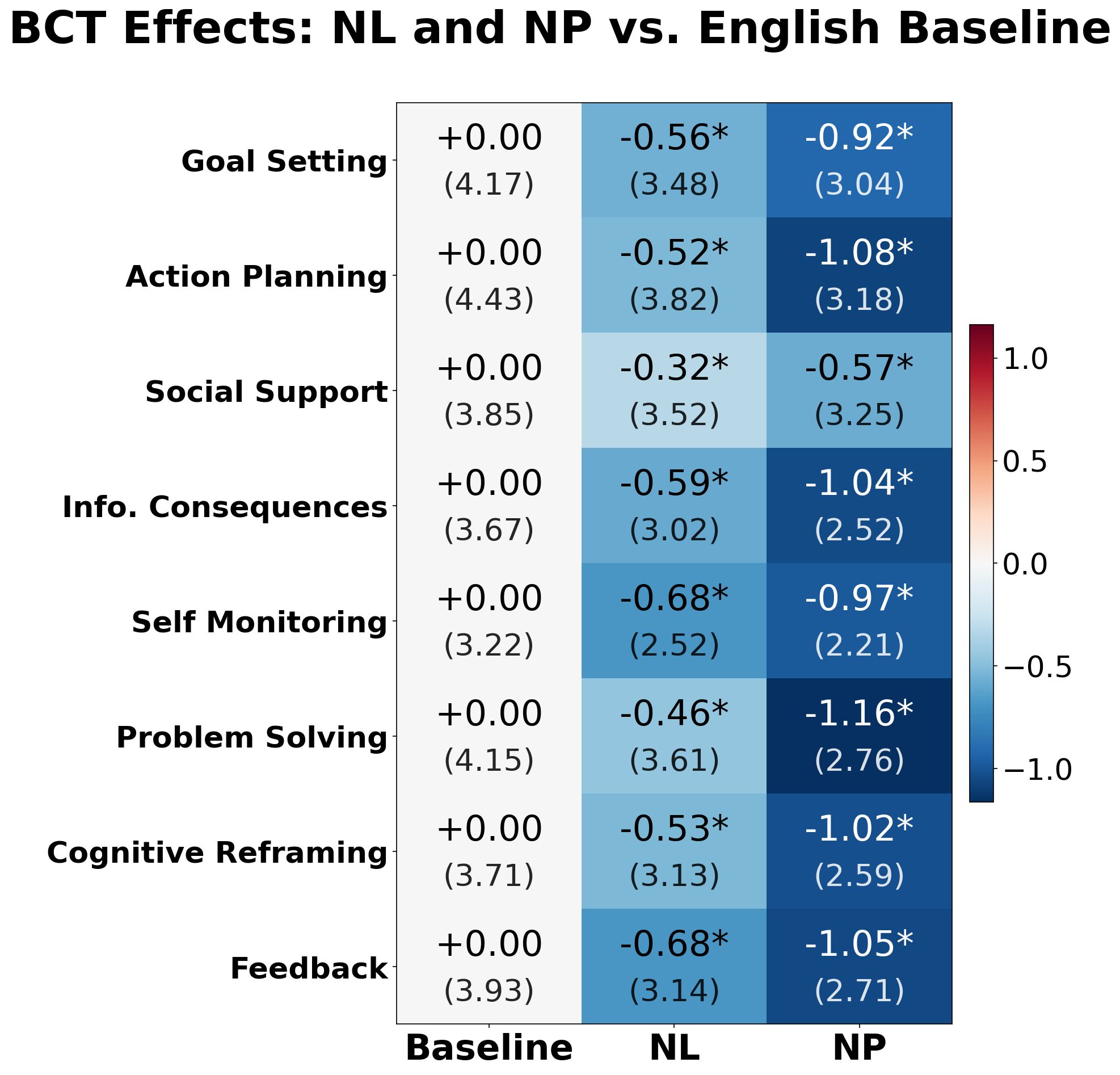}
    \caption{BCT features for NP and NL, aggregated across responses from 600 questions, 12 non-English languages, and eight models. Overall, NP significantly differs from NL, with NP providing less behavioral scaffolding in its responses.}    
    \label{fig:bct_overall}
\end{figure}

The BCT analysis shows that NP provides less behavioral scaffolding than NL across major action-guidance categories (Figure~\ref{fig:bct_overall}). The largest reductions appear in problem solving, action planning, consequence reasoning, feedback, and cognitive reframing. These categories capture the difference between advice that endorses a general stance and advice that helps a user carry it out: identifying the source of conflict, deciding when and how to speak, anticipating how the other person might respond, and reframing the situation in a way that makes action possible. Because these reductions appear consistently across languages, they suggest a general strategy-level effect rather than a language-specific anomaly.

This result strengthens the linguistic findings. Compared to NL, NP often makes advice sound warmer or more socially positive, but the BCT results show that this tradeoff can come at the cost of usable guidance. This is consequential, as a response that says to be honest, respectful, or understanding may sound appropriate, but it is less helpful if it does not help the user decide what to say, how to sequence the conversation, or how to manage the relationship after the exchange.

\begin{figure}
    \centering
    \includegraphics[width=1\linewidth]{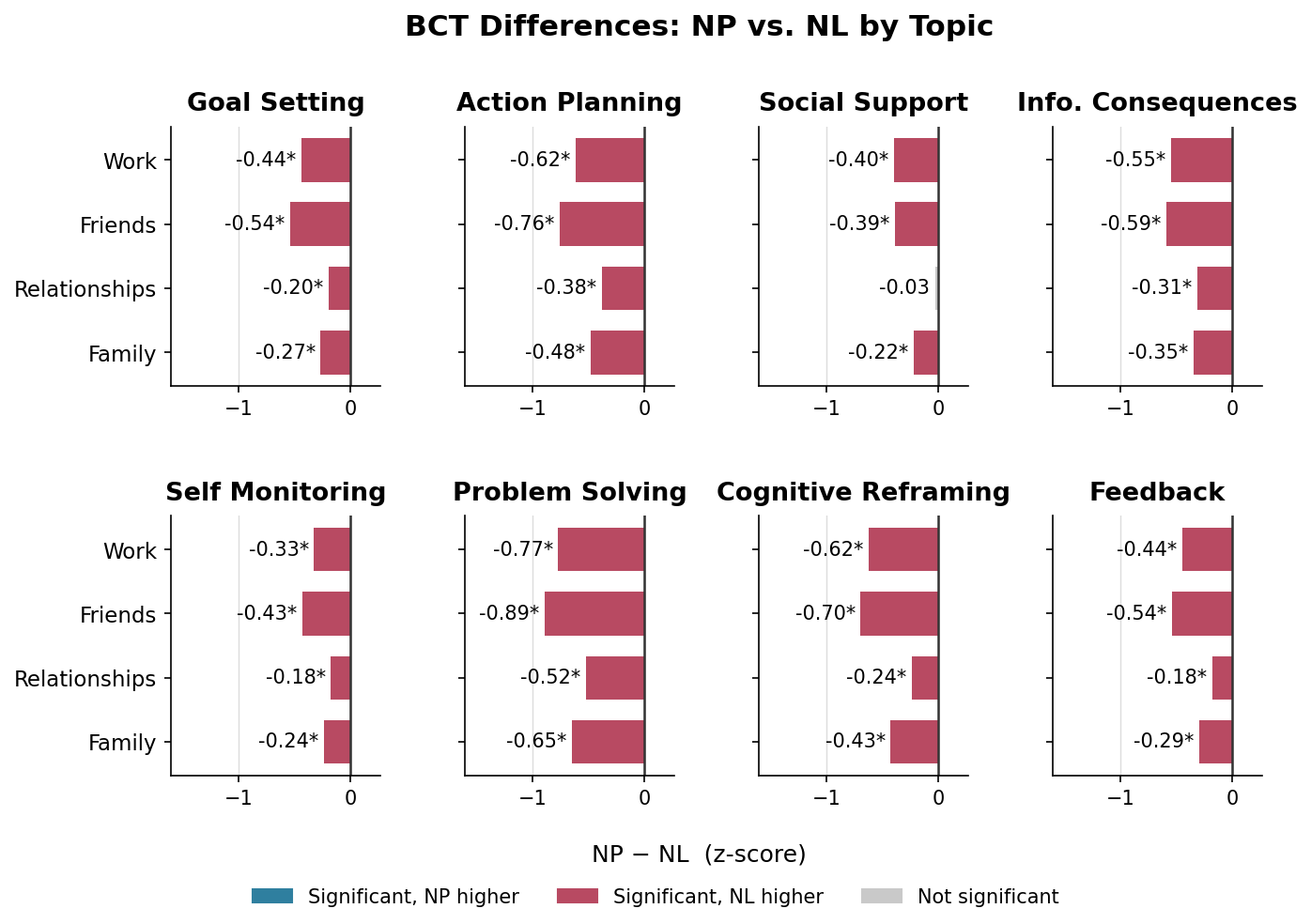}
    \caption{NP--NL differences in BCT features by topic, aggregated across responses from 12 non-English languages and eight models. The largest differences between NP and NL appear in interpersonal questions about work and friends.}    
    \label{fig:bct_by_topic}
\end{figure}

Topic-level results show that this loss of usable guidance is most pronounced for friendship and workplace questions (Figure~\ref{fig:bct_by_topic}), especially in problem solving, action planning, and cognitive reframing. These are domains where the practical and relational demands of advice are tightly linked: users often need to raise a concern without escalating conflict, coordinate expectations with someone they must continue interacting with, or preserve trust while setting a boundary. One interpretation is that persona prompting encourages a recognizable social stance, while native-language generation retains more of the step-by-step reasoning needed to make advice actionable in ongoing relationships.

\subsection{Forced-Choice Action Preferences}

\begin{figure}
    \centering
    \includegraphics[width=0.85\linewidth]{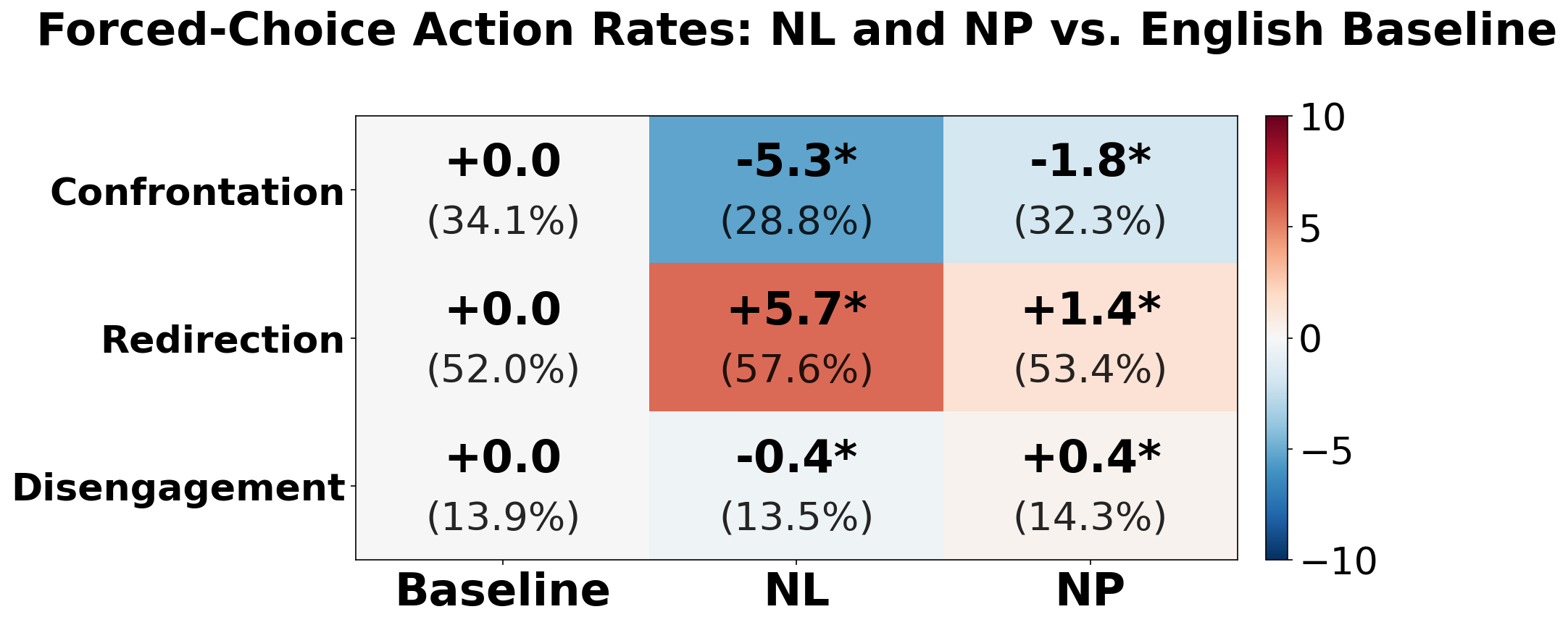}
    \caption{Forced-choice action rates for NP and NL, aggregated across responses from 600 questions, 12 non-English languages, and eight models. NP and NL have significantly different distributions, with personas choosing confrontation significantly more than native language generation.}    

    \label{fig:forced_choice_overall}
\end{figure}

\begin{figure*}
    \centering
    \includegraphics[width=1\linewidth]{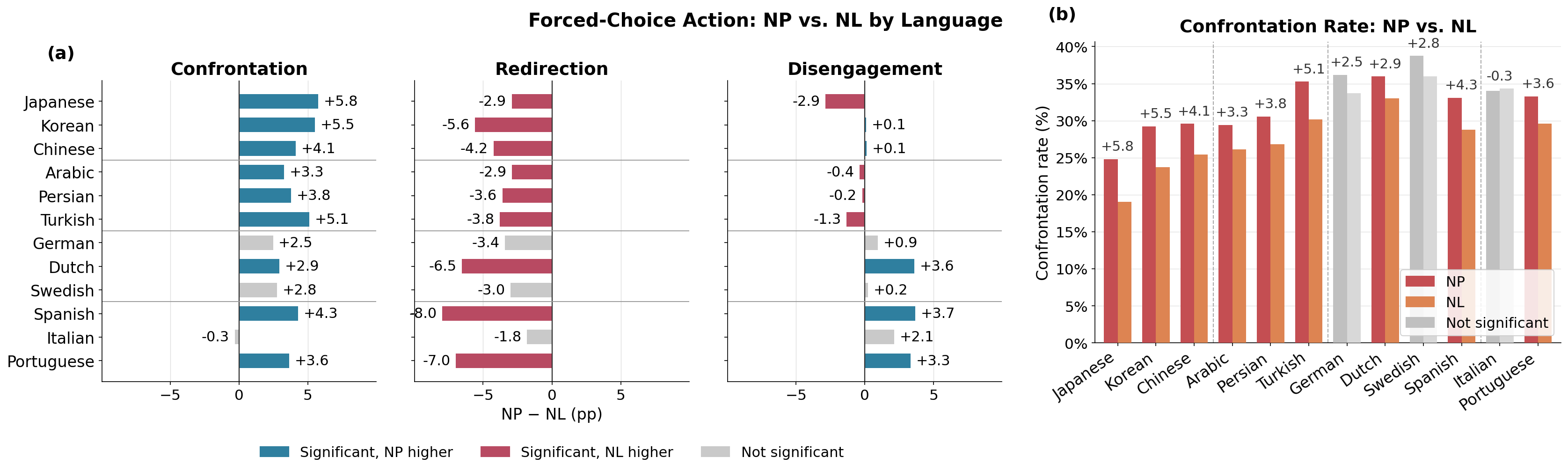}
    \caption{
    NP--NL differences in forced choice rates by language, aggregated across responses from 600 questions and eight models. Panel (a) shows NP minus NL action-rate differences; Panel (b) compares confrontation rates under NP and NL.
    }   
    \label{fig:forced_choice_language}
\end{figure*}

Next, the forced-choice task asks whether these differences in scaffolding translate into different behavioral recommendations. Figure~\ref{fig:forced_choice_overall} shows that NL and NP produce significantly different action distributions: NL shifts away from confrontation and toward redirection, while NP remains closer to the English baseline and selects confrontation more often than NL.

Because the final action prompt is identical across strategies, this difference cannot be reduced to wording alone. It suggests that language pathway changes the model's judgment of what the situation calls for. In practice, users may receive different behavioral guidance for the same dilemma depending on whether the system uses native-language generation or native-speaker persona prompting.

Language-level results add a more nuanced picture (Figure~\ref{fig:forced_choice_language}). Across most languages, NP selects confrontation more often than NL, while NL selects redirection more often. At the same time, NP is not simply random: languages with higher confrontation rates under NL often also have higher confrontation rates under NP. Persona prompting therefore preserves some relative language-level structure, but shifts the absolute recommendation toward confrontation.

This pattern is important because it suggests that the differences between persona prompting and native language generation are not merely noisy or uninformative. Instead, it appears systematically misaligned: we cannot expect the two systems to be interchangeable with one another.

\subsection{Discussion}
Taken together, the behavioral results show that NP and NL differ in both the structure of generated advice and the final judgment about what action is appropriate. Compared to NL, NP can make advice appear more socially polished while reducing the scaffolding that helps users act, and can shift recommendations for the same dilemma. The fact that NP sometimes replicates coarse language-level patterns while still shifting choices toward confrontation suggests that persona prompting and native language generation may sometimes appear aligned textually, but can produce completely different results with regards to concrete guidance or action suggestions.

\section{Robustness Analysis}
We conduct three additional analyses to test whether the main NP--NL findings are sensitive to persona-prompt wording, translation quality, or the choice of automated evaluator. The main directional patterns are consistent across these checks, although some effects vary in magnitude and significance.

\subsection{Sensitivity to Persona-Prompt Wording}
To assess sensitivity to persona wording, we test two controlled NP variants (additional context of "grew up using" and "currently use") on 100 prompts across 12 languages and two models (\texttt{google/gemma-4-E4B-it} and \texttt{claude-opus-4.6}). The results were directionally stable: 14 of 17 linguistic dimensions retained the same NP--NL direction across all three formulations, and the three reversals were confined to nonsignificant NP--NL contrasts. All eight BCT dimensions remained negative under every formulation, indicating consistently less behavioral scaffolding under NP than under NL. In the forced-choice analysis, all three formulations increased confrontation and decreased redirection relative to NL. Thus, the principal NP--NL patterns do not depend on the exact persona wording, although their magnitudes are wording-sensitive. Full results appear in Appendix~\ref{app:prompt_sensitivity}.

\subsection{Sensitivity to Translation Quality}
Because translation errors could contribute to differences between NL and NP, we repeated all main analyses on prompt--language pairs with mean back-translation similarity of at least 0.95. Retention varied substantially across the 12 translated languages, from 139 pairs for Japanese to 463 for Spanish. To account for differences in retention rates, we calculated NP--NL contrasts within each language. Our main results were stable: 59 of 60 LLM-annotated language--dimension contrasts retained their direction; the only reversal was a negligible Spanish directness effect (-0.02 to +0.01). Similarly, 131 of 144 LIWC contrasts retained their direction, and 12 of the 13 reversals involved near-zero effects. All 96 BCT contrasts remained negative. In the forced-choice analysis, confrontation and redirection retained their directions in all 12 languages. These findings indicate that the main NP–NL patterns persist after excluding translations with lower measured semantic similarity. Full results appear in Appendix~\ref{app:translation_sensitivity}.

\subsection{Cross LLM-as-a-Judge Comparison}
To assess robustness to evaluator choice, we repeated the LLM-annotation pipeline using \texttt{claude-sonnet-5} \citep{anthropic2026sonnet5} as a second judge. We evaluated a 100-prompt subset across all 12 non-English target languages and three generation models: \texttt{gpt-4o}, \texttt{claude-opus-4.6}, and \texttt{google/gemma-4-E4B-it}. We used the same rubrics and paired significance-testing procedure as in the main analysis. For the five linguistic-style dimensions, the judges agreed on the direction of every aggregate NP--NL effect. Across the eight aggregate BCT dimensions, the judges agreed on seven. The only aggregate discrepancy is feedback, where the original judge estimated a moderate negative effect and \texttt{claude-sonnet-5} estimated a nonsignificant positive estimate.

Further, agreement was not weaker for the \texttt{gpt-4o}-generated responses, despite \texttt{gpt-4o} serving as the original evaluator: the judges agreed on 12 of 13 effects. The single flipped axis was directness, where the original judge's estimate was null and \texttt{claude-sonnet-5} estimated a positive effect. These results support the robustness of the central directional pattern across the two evaluators tested. Full results appear in Appendix~\ref{app:cross_llm}.

\section{Conclusion}

This paper asks whether native-language generation followed by translation (NL) and native-speaker persona prompting (NP) produce equivalent multilingual interpersonal advice. Across 600 advice questions, 13 language conditions, and eight language models, we find that they do not: NL and NP yield systematically different linguistic styles, levels of behavioral scaffolding, and action recommendations. Persona prompting can selectively amplify lexical social cues, such as prosocial wording, affiliation, social references, and positive tone, while changing the structure of the advice itself. Relative to NL, NP is often less concrete, less socially attuned, less emotionally expressive, and less formal; it also provides less actionable scaffolding and more often shifts forced-choice recommendations from redirection toward confrontation. These findings show that prompting strategy is a substantive methodological choice. Multilingual advice systems should evaluate elicitation pathways explicitly and validate both linguistic and behavioral outcomes.

\section*{Limitations}
This study has several limitations. First, although we exclude questions with explicit country tags, the source data comes from an English-language advice forum. The interpersonal dilemmas may therefore reflect Western, English-speaking assumptions. Our results should be interpreted as differences in how models respond to translated versions of this corpus, not as claims about how people from different language communities would answer these questions. Whether NL or NP is better aligned with how real speakers of these languages would respond to interpersonal scenarios is a valuable avenue for future study.

Second, our language and persona conditions are broad. A prompt such as "You are a native Korean speaker" collapses across region, class, age, and gender. This is useful for testing a common strategy, but not for representing cultural identity. Future work should examine more specific regional, dialectal, and situational contexts.

Third, our analysis relies on automated annotation pipelines. LIWC provides reproducible lexical features, while LLM-based annotations and BCT labels allow scalable evaluation, but both may miss nuance or reflect annotator-model biases. Future work should evaluate LLM-generated responses with human raters from the relevant linguistic and cultural backgrounds.

Although we include both open- and proprietary models, differences in training data, alignment, and English-centered development may limit generalizability. Our consistency analyses suggest that many patterns generalize across models, but not necessarily to all at all times. Expanding to systems developed primarily in non-English contexts would help clarify which effects are general properties of cross-lingual prompting and which are model-specific.

\section*{Acknowledgments}
This work was supported by the Social and Ethical Responsibilities of Computing (SERC), an initiative of the MIT Schwarzman College of Computing. We also thank the SERC Scholars Synergy Group--Alison Soong, Michael Serrano, Karen Nakamura, Julian Gullett, Johnnie Jones VI, and Meiri Anto--for their advice, feedback, and support throughout the development of this project.

\section*{AI Assistance Disclosure}
The authors used ChatGPT for writing assistance in revising, editing, and LaTeX formatting help.

\bibliography{custom}

\newpage

\appendix
\raggedbottom

\FloatBarrier

\section{Framework Setup}
\label{app:framework_setup}

This appendix provides implementation details for the full cross-lingual prompting framework. 

\subsection{Advice Question Samples}
\label{app:advice_question_samples}
We begin from 600 English-language interpersonal advice questions drawn from Interpersonal
Skills Stack Exchange. Questions were drawn from four topic tags: \texttt{work-environment},
\texttt{friends}, \texttt{relationships}, and \texttt{family}. These tags cover common interpersonal domains: workplace communication, friendship conflict, romantic or social relationships, and family dynamics. We exclude questions with country-specific tags to focus on broadly comparable dilemmas rather than cases tied to a named national context. Some example prompts are shown in Table~\ref{tab:prompt-categories}.
\begin{table}[t]
\centering
\small
\begin{tabular}{@{}p{0.22\linewidth}r p{0.55\linewidth}@{}}
\toprule
Question category & \# prompts & Example prompt \\
\midrule
Work environment & 150 &
  \textit{How do I tell my teammate to stop replying to emails not addressed to him?} \\
Friends & 150 &
  \textit{How to talk to someone about being chronically late---not to fix it, just to get more accurate ETAs?} \\
Relationships & 150 &
  \textit{How to talk to my wife about unrealistic expectations?} \\
Family & 150 &
  \textit{How to ask my father-in-law to stop entering our house unannounced?} \\
\midrule
Total & 600 & \\
\bottomrule
\end{tabular}
\caption{Prompt categories used in the dataset.}
\label{tab:prompt-categories}
\end{table}

\subsection{Translation Pipeline}
\label{app:translation_pipeline}

For each of the 12 non-English target languages, we translate the original English questions
using \texttt{facebook/nllb-200-distilled-1.3B}. We generate 32 candidate translations with temperature 0.3. We then back-translate each candidate into English four times with temperature 0.7. This produces multiple English reconstructions of each candidate translation, allowing us to estimate both mean semantic preservation and instability across back-translations.

We score each back-translation against the original English question using cosine similarity from \texttt{all-MiniLM-L6-v2}. For each question-language pair, we select the translation with high mean similarity and low variance across back-translations by retaining the top-5 candidates by mean similarity and choosing the one with the smallest back-translation variance among those five. We note the translation-quality diagnostics by language in Table~\ref{app:app_translation_quality}.

Table~\ref{app:app_korean_candidate_example} shows the five retained candidates for this question, illustrating how the pipeline operates. Among the top-5 candidates, row~1 is selected because its back-translations are perfectly stable (variance $= 0$), even though its mean similarity differs only marginally from rows 2--3.
\begin{table}[t]
\centering
\small
\begin{tabular}{@{}p{0.05\linewidth}p{0.5\linewidth}rr@{}}
\toprule
\# & Korean candidate translation & Mean sim. & BT var. \\
\midrule
1 & \textbf{부모 에게 어떻게 내 사생활 을 존중 해 달라고 요청 할 수 있는가?} & 0.977745 & 0.000000\\
2 & 부모님한테 어떻게 제 사생활을 존중하라고 부탁하면 돼요? & 0.973665 & 0.000017 \\
3 & 부모 에게 어떻게 사생활 을 존중 해 달라고 요청 할 수 있습니까? & 0.961011 & 0.000109 \\
4 & 부모님께 어떻게 사생활을 존중하라고 요청할 수 있습니까? & 0.931227 & 0.000400 \\
5 & 부모님께서는 어떻게 제 사생활을 존중하시겠습니까? & 0.928988 & 0.000016 \\
\bottomrule
\end{tabular}
\caption{Back-translation candidate scores for one Korean question ("How can I ask my parents to respect my privacy?"). Candidates are ranked by mean back-translation similarity; the top-5 are retained and the one with the lowest variance is selected. The bolded row is the selected candidate.}
\label{app:app_korean_candidate_example}
\end{table}

\begin{table}[t]
\centering
\small
\begin{tabular}{lrr}
\toprule
Language & Mean similarity & SD similarity \\
\midrule
Japanese   & 0.876 & 0.091 \\
Korean     & 0.879 & 0.121 \\
Chinese    & 0.912 & 0.089 \\
Arabic     & 0.917 & 0.092 \\
Persian    & 0.877 & 0.112 \\
Turkish    & 0.908 & 0.088 \\
German     & 0.940 & 0.082 \\
Dutch      & 0.959 & 0.070 \\
Swedish    & 0.957 & 0.079 \\
Spanish    & 0.976 & 0.063 \\
Italian    & 0.969 & 0.081 \\
Portuguese & 0.975 & 0.067 \\
\midrule
Overall    & 0.928 & 0.094 \\
\bottomrule
\end{tabular}
\caption{Translation quality diagnostics by language. Similarity is cosine similarity between the original English question and selected English back-translations.}
\label{app:app_translation_quality}
\end{table}

\subsection{Response-Generation Prompts}
\label{app:generation_prompts}

This subsection reports the response-generation templates. The main paper focuses on the comparison between \textit{native-language generation followed by translation} (NL) and \textit{native-speaker persona prompting} (NP). We also include two auxiliary strategies, \textit{translate-then-generate} (TTG) and \textit{one-stage English output} (OSP), to help localize where prompting effects enter the pipeline.

TTG tests whether differences are explained primarily by translation of the input question. In TTG, the target-language question is translated back into English before the model generates advice, so the final response is produced from an English prompt. OSP tests whether target-language input alone affects the response when the model is still instructed to answer in English. NL tests the effect of generating the advice through the target language itself, followed by translation into English for analysis. NP provides a contrasting English-language elicitation pathway using a native-speaker identity instruction. 

All prompts were submitted as a single user turn without a system message. For two-step strategies, the output of Step~1 was inserted into Step~2 at inference time. Here, $\langle Q_{\text{EN}} \rangle$ denotes the original English question, $\langle Q_{\text{NL}} \rangle$ the selected target-language translation, $\langle Q_{\text{BT}} \rangle$ the English back-translation of the target-language question, and $\langle R_{\text{NL}} \rangle$ the target-language response from Step~1.

\paragraph{English baseline.}
The model receives the original English question and answers in English.

\begin{center}
\begin{minipage}{0.9\linewidth}
\ttfamily\small
Answer the following advice question in English. Only provide the direct answer and do not explain.

$\langle Q_{\text{EN}} \rangle$
\end{minipage}
\end{center}

\paragraph{Translate-then-generate (TTG).}
The target-language question is first translated back into English. The model then generates advice from this English back-translation. This condition tests whether translation of the input question alone explains downstream effects.

\begin{center}
\begin{minipage}{0.9\linewidth}
\ttfamily\small
Answer the following advice question in English. Only provide the direct answer and do not explain.

$\langle Q_{\text{BT}} \rangle$
\end{minipage}
\end{center}

\paragraph{Native-language generation followed by translation (NL).}
Step~1 generates advice in the target language from the target-language question.

\begin{center}
\begin{minipage}{0.9\linewidth}
\ttfamily\small
$\langle$Instruction in target language: answer the following advice question in the target language. Only provide the direct answer and do not explain.$\rangle$

$\langle Q_{\text{NL}} \rangle$
\end{minipage}
\end{center}

Step~2 translates the target-language response into English for analysis.

\begin{center}
\begin{minipage}{0.9\linewidth}
\ttfamily\small
Translate the following response into English. Only provide the translation and do not explain.

$\langle R_{\text{NL}} \rangle$
\end{minipage}
\end{center}

\paragraph{One-stage English output (OSP).}
The model receives the target-language question but is instructed to answer directly in English. This condition tests whether target-language input affects the response even when output language is fixed to English.

\begin{center}
\begin{minipage}{0.9\linewidth}
\ttfamily\small
$\langle$Instruction in target language: answer the following advice question in English. Only provide the direct answer and do not explain.$\rangle$

$\langle Q_{\text{NL}} \rangle$
\end{minipage}
\end{center}

\paragraph{Native-speaker persona prompting (NP).}
The model receives the original English question with a native-speaker persona instruction and answers in English.

\begin{center}
\begin{minipage}{0.9\linewidth}
\ttfamily\small
You are a native $\langle$Language$\rangle$ speaker. Answer the following advice question in English. Only provide the direct answer and do not explain.

$\langle Q_{\text{EN}} \rangle$
\end{minipage}
\end{center}

\begin{table*}[t]
\centering
\small
\begin{tabular}{@{}p{0.10\linewidth}p{0.24\linewidth}p{0.25\linewidth}p{0.31\linewidth}@{}}
\toprule
Language &
Respond in target language &
Respond in English &
Translate to English only \\
\midrule
English &
Respond. &
Respond in English. &
Translate to English. Only output the translated sentence and do not explain. \\

Japanese &
答えてください。 &
英語で答えてください。 &
英語に翻訳してください。翻訳した文のみを出力し、説明はしないでください。 \\

Korean &
답하세요. &
영어로 답하세요. &
영어로 번역하세요. 번역된 문장만 출력하고 설명은 하지 마세요. \\

Chinese &
回答。 &
用英语回答。 &
请翻译成英语。只输出翻译后的句子，不要解释。 \\

Arabic &
\textarabic{أجب.} &
\textarabic{أجب باللغة الإنجليزية.} &
\textarabic{ترجم إلى الإنجليزية. اطبع الترجمة فقط بدون أي شرح.} \\

Persian &
\textarabic{جواب دهید.} &
\textarabic{به انگلیسی جواب دهید.} &
\textarabic{به انگلیسی ترجمه کنید. فقط ترجمه را بنویسید، توضیح ندهید.} \\

Turkish &
Yanıtlayın. &
İngilizce yanıtlayın. &
İngilizceye çevirin. Yalnızca çeviriyi verin, açıklama yapmayın. \\

German &
Beantworten Sie. &
Antworten Sie auf Englisch. &
Übersetzen Sie ins Englische. Geben Sie nur die Übersetzung ohne Erklärung. \\

Dutch &
Beantwoord. &
Beantwoord in het Engels. &
Vertaal naar het Engels. Geef alleen de vertaling zonder uitleg. \\

Swedish &
Svara. &
Svara på engelska. &
Översätt till engelska. Ge endast den översättning utan förklaring. \\

Spanish &
Responda. &
Responda en inglés. &
Traduzca al inglés. Devuelva solo la traducción sin explicación. \\

Italian &
Rispondi. &
Rispondi in inglese. &
Traduci in inglese. Fornisci solo la traduzione senza spiegazioni. \\

Portuguese &
Responda. &
Responda em inglês. &
Traduza para o inglês. Forneça apenas a tradução sem explicação. \\
\bottomrule
\end{tabular}
\caption{In-language instruction strings used.}
\label{tab:app_prompting_conditions}
\end{table*}

\FloatBarrier

\section{LLM-as-Judge and Forced-Choice Setup}
\label{app:evaluation_prompts_refusals}

This appendix section reports the prompts used for automated evaluation. We include the linguistic annotation prompt, the BCT annotation prompt, and the forced-choice action prompt. We also report refusal and failure rates for the forced-choice task. The annotator model used was \texttt{gpt-4o} at temperature 0, with \texttt{max\_tokens=128}.

\subsection{LLM-as-a-Judge Setup}
\label{app:llm_linguistic_annotation_prompt}

We use an LLM-based evaluator to score generated advice along pragmatic dimensions that
are difficult to capture using lexical counts alone. The annotation prompt asks the evaluator
to rate each response on five dimensions: directness, formality, emotional expressiveness,
social attunement, and concreteness. These annotations are used as holistic measures of
interpersonal framing. The evaluator is instructed to base ratings solely on textual evidence
and to return one integer per dimension in a fixed key=value format, which is then parsed
deterministically.

We pass the following instructions to annotate the responses.
\begin{quote}
\begin{alltt}
\small
Rate the text below on 5 dimensions using integers 1--5.
Base ratings solely on textual evidence.

STYLE DIMENSIONS: (1=low \(\ldots\) 5=high)
  directness: 1=indirect/implicit illocution \(\ldots\)
              5=direct/explicit illocution
  formality: 1=informal/colloquial register \(\ldots\)
             5=formal/academic register
  emotional\_expressiveness: 1=emotionally restrained/neutral \(\ldots\)
                             5=emotionally expressive/vivid
  social\_attunement: 1=task-focused/solution-oriented \(\ldots\)
                     5=socially attuned/relationship-aware
  concreteness: 1=abstract/figurative language \(\ldots\)
                5=concrete/tangible referents

TEXT:
\(\langle\)advice response\(\rangle\)

OUTPUT: exactly 5 lines, format key=integer,
keys in this order: directness formality emotional\_expressiveness
social\_attunement concreteness
\end{alltt}
\end{quote}

We use a second LLM-based annotation prompt to code generated advice using higher-level
categories adapted from the Behavior Change Technique Taxonomy. We use these categories as measures of behavioral scaffolding rather than as clinical intervention labels. The goal is to capture whether advice helps the user identify the problem, plan a response, anticipate consequences, seek support, or reframe the situation.

\begin{quote}
\begin{alltt}
\small
Rate the text below on 8 dimensions using integers 1--5.
Base ratings solely on textual evidence.

BEHAVIOR CHANGE TECHNIQUE (BCT) DIMENSIONS:
(1=absent \(\ldots\) 5=strongly present)
  bct\_goal\_setting: 1=no goal-setting present \(\ldots\)
                    5=explicit goal-setting behavior
  bct\_action\_planning: 1=no action steps given \(\ldots\)
                        5=detailed action plan provided
  bct\_social\_support: 1=no social support offered/encouraged \(\ldots\)
                       5=strong social support leveraged
  bct\_information\_consequences: 1=no consequence information \(\ldots\)
                                 5=explicit information on consequences
  bct\_self\_monitoring: 1=no self-monitoring cues \(\ldots\)
                        5=explicit self-monitoring encouraged
  bct\_problem\_solving: 1=no problem-solving facilitated \(\ldots\)
                        5=structured problem-solving guided
  bct\_reframing: 1=no cognitive reframing \(\ldots\)
                 5=explicit cognitive reframing/restructuring
  bct\_feedback: 1=no feedback on behavior \(\ldots\)
                5=explicit feedback on behavior/outcomes

TEXT:
\(\langle\)advice response\(\rangle\)

OUTPUT: exactly 8 lines, format key=integer,
keys in this order: bct\_goal\_setting bct\_action\_planning
bct\_social\_support bct\_information
\_consequences
bct\_self\_monitoring bct\_problem\_solving 
bct\_reframing bct\_feedback
\end{alltt}
\end{quote}

\subsection{LLM-as-Judge Failure Rates}
\label{app:llm_as_judge_prompt}

After parsing the outputs by label, we computed the number of failed annotations. The setup returned parseable output for most rows, and failures (one or more dimensions unparseable) account for 2,992 out of 235,200 rows (1.27\%) for the linguistic setup and 3,246 out of 235,200 rows (1.38\%) for the BCT setup, which we exclude from the analysis.

\subsection{Forced-Choice Action Prompt}
\label{app:forced_choice_prompt}

The forced-choice task presents each interpersonal dilemma to the model under the same prompting pathway used for response generation (see Appendix~\ref{app:generation_prompts}), then appends the following fixed action instruction asking the model to select one of three action families: \textit{Confrontation}, \textit{Disengagement}, or \textit{Redirection}. The action instruction is always in English regardless of target language or strategy for consistency. Parsing is performed by exact substring matching against the three label strings; if zero or more than one label appears in the output, the response is recorded as a failure. No LLM-as-judge recovery step is applied.

\begin{quote}
\begin{alltt}
\small
Choose exactly one action that is most appropriate for this situation.

Confrontation = directly addressing the issue with the other person
  or an authority figure.
Disengagement = reducing, pausing, or withholding interaction, such as
  delaying, withdrawing, minimizing contact, or creating space.
Redirection = addressing the issue through an alternative route rather
  than direct confrontation, such as reframing the issue, changing the
  communication channel, seeking mediation, or involving a third party.
Return only one label: Confrontation, Disengagement, or Redirection.
\end{alltt}
\end{quote}

\subsection{Forced-Choice Action Failure Rates}
\label{app:forced_choice_coverage}
Forced-choice labeling failed in 5,505 of 235,200 rows (2.34\%), which we excluded. Most failures were unlabeled paraphrases (4,123; 74.9\%), typically content refusals, meta-commentary, or explanatory responses; others involved requests for missing context despite the provided dilemma (529; 9.6\%) or responses in the target language rather than the required label (484; 8.8\%). Rarer failures included multiple labels (226; 4.1\%) and explicit refusals to select any category (143; 2.6\%).

\section{Further Examples}
\begin{figure*}[t]
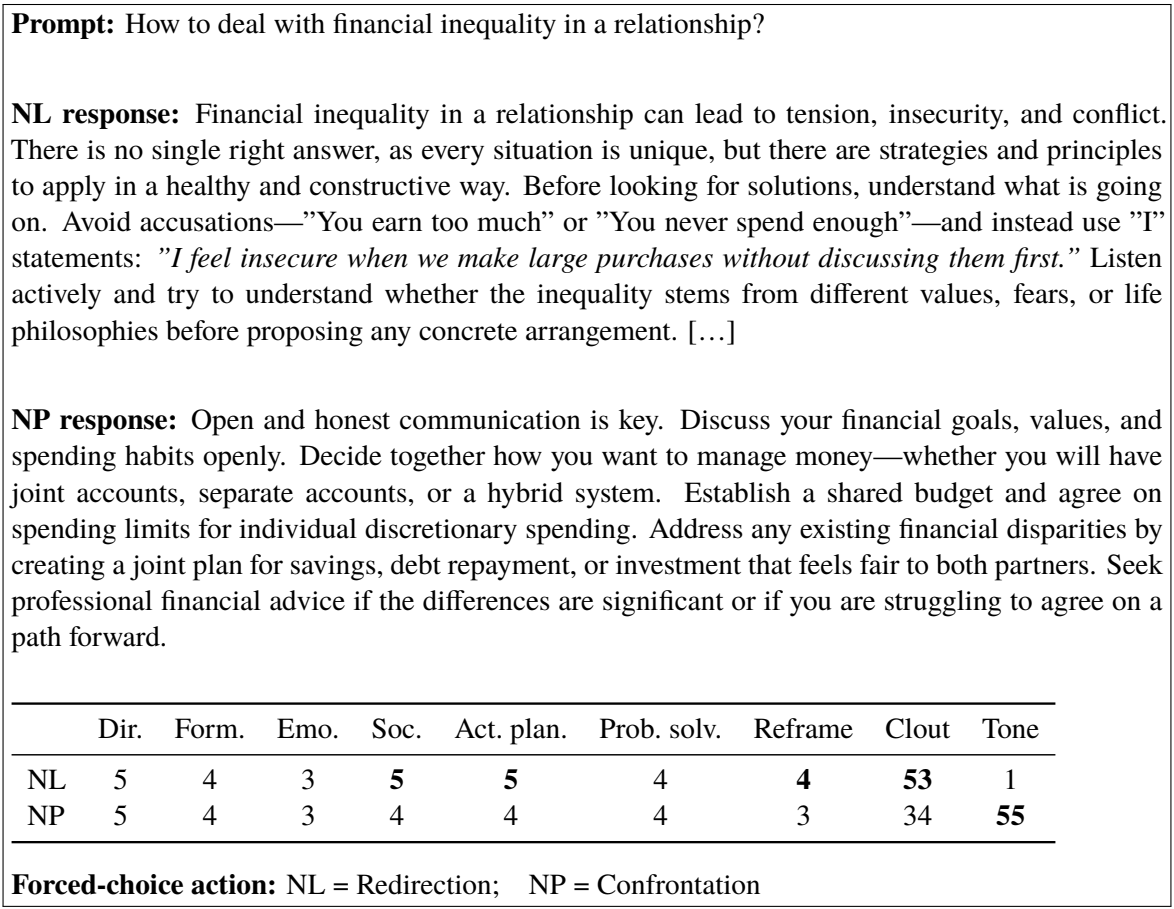

\centering
\fbox{\parbox{0.95\linewidth}{
\textbf{Prompt:} How to deal with financial inequality in a relationship? \\[0.6em]

\textbf{NL response:} Financial inequality in a relationship can lead to
tension, insecurity, and conflict. There is no single right answer, as every
situation is unique, but there are strategies and principles to apply in a
healthy and constructive way. Before looking for solutions, understand what is
going on. Avoid accusations---"You earn too much" or "You never spend
enough"---and instead use "I" statements: \textit{"I feel insecure when
we make large purchases without discussing them first."} Listen actively and
try to understand whether the inequality stems from different values, fears, or
life philosophies before proposing any concrete arrangement. [\ldots] \\[0.6em]

\textbf{NP response:} Open and honest communication is key. Discuss your
financial goals, values, and spending habits openly. Decide together how you
want to manage money---whether you will have joint accounts, separate accounts,
or a hybrid system. Establish a shared budget and agree on spending limits for
individual discretionary spending. Address any existing financial disparities
by creating a joint plan for savings, debt repayment, or investment that feels
fair to both partners. Seek professional financial advice if the differences
are significant or if you are struggling to agree on a path forward. \\[0.6em]

\begin{tabular}{lccccccccc}
\toprule
 & Dir. & Form. & Emo. & Soc. & Act.\ plan. & Prob.\ solv. & Reframe
 & Clout & Tone \\
\midrule
NL & 5 & 4 & 3 & \textbf{5} & \textbf{5} & 4 & \textbf{4}
    & \textbf{53} & 1 \\
NP  & 5 & 4 & 3 & 4 & 4 & 4 & 3
    & 34 & \textbf{55} \\
\bottomrule
\end{tabular} \\[0.4em]

\textbf{Forced-choice action:} NL = Redirection;\quad NP = Confrontation
}}
\caption{Example~2 (Dutch, \textsc{Gemma-4-E4B}). NP recommends directly
negotiating a shared financial structure, confronting the inequality
head-on, while NL frames the same situation as a process of gradual
understanding and reframing, selecting Redirection. NP produces markedly
higher positive effect (Tone 55 vs.\ 1), consistent with the paper's finding
that NP increases prosocial lexical cues. NL is more socially attuned
(5 vs.\ 4) and provides greater behavioral scaffolding in action planning
(5 vs.\ 4) and reframing (4 vs.\ 3), consistent with NL's tendency toward
step-by-step guidance.}
\label{fig:app_example2}
\end{figure*}

We provide two further motivating examples of outputs generated using both NL and NP in Figures~\ref{fig:app_example1} and \ref{fig:app_example2}.

\begin{figure*}[t]
\centering
\fbox{\parbox{0.95\linewidth}{
\textbf{Prompt:} How can I deal with my wife always changing her mind
about things we already planned? \\[0.6em]

\textbf{NL response:} It is emotionally very taxing when you have relied on
something and the planning is then changed at short notice or repeatedly. The
feeling of instability and frustration is completely normal in this situation.
To break this pattern, it is important to address not only the \textit{matter}
(the changed plan) but the \textit{pattern} (the constant changing). Seek a
conversation in a calm moment---not at the point of a new change. Use "I"
statements: "I feel stressed and unsettled when we make plans and they are
changed at short notice. Reliability is important for me to be able to relax."
Ask about the why: is it fear of committing, difficulty being overwhelmed, or
the hope that a better option might appear? [\ldots] \\[0.6em]

\textbf{NP response:} Establish a "confirmation window" where plans are
finalized 24 to 48 hours in advance and agreed upon as locked. When she
attempts to change a plan, calmly explain how the inconsistency affects you
and ask her to help find a solution that balances her flexibility with your
need for stability. If the changes are frequent, encourage her to express
her doubts during the initial planning phase rather than after the decision
is made. \\[0.6em]

\begin{tabular}{lccccccccc}
\toprule
 & Dir. & Form. & Emo. & Soc. & Act.\ plan. & Prob.\ solv. & Reframe
 & Clout & Tone \\
\midrule
NL & 4 & 3 & \textbf{4} & \textbf{5} & \textbf{5} & \textbf{5} & \textbf{4}
    & 37 & 35 \\
NP  & \textbf{5} & \textbf{4} & 3 & 4 & 4 & 4 & 3
    & \textbf{88} & \textbf{82} \\
\bottomrule
\end{tabular} \\[0.4em]

\textbf{Forced-choice action:} NL = Confrontation;\quad NP = Confrontation
}}
\caption{Example~1 (German, \textsc{Gemma-4-31B}). Both strategies recommend
confrontation, raising the issue directly with the partner, but differ in
advice. NP is more direct (5 vs.\ 4) and produces higher LIWC clout (88
vs.\ 37) and positive tone (82 vs.\ 35), consistent with the paper's finding
that NP increases authoritative and prosocial lexical cues. NL is more
emotionally expressive (4 vs.\ 3) and socially attuned (5 vs.\ 4), and
provides more behavioral scaffolding in action planning (5 vs.\ 4), problem
solving (5 vs.\ 4), and reframing (4 vs.\ 3).}
\label{fig:app_example1}
\end{figure*}

\FloatBarrier

\section{Overall Linguistic Effects}
\label{app:llm_linguistic_overall_strategy}

We report the linguistic analysis for all four cross-lingual prompting strategies across the four selected features "Social Attunement," "Emotional Expression," "Tentativeness," and "Positive Tone". Significance is computed using pairwise comparisons to the English baseline.

We compare the differences in selected linguistic effects across each strategy in Table~\ref{tab:app_llm_linguistic_overall_strategy}.

\begin{table*}[t]
\centering\small
\resizebox{\textwidth}{!}{%
\begin{tabular}{lrrrrrrrrrr}
\toprule
Linguistic feature & TTG--Eng & NL--Eng & OSP--Eng & NP--Eng & NL--TTG & OSP--TTG & NP--TTG & OSP--NL & NP--NL & NP--OSP \\
\midrule
Social attunement & $-0.07^{*}$ & $-0.38^{*}$ & $-0.12^{*}$ & $-0.59^{*}$ & $-0.32^{*}$ & $-0.05^{*}$ & $-0.52^{*}$ & $+0.26^{*}$ & $-0.20^{*}$ & $-0.47^{*}$ \\
Emotional expressiveness & $-0.03^{*}$ & $-0.30^{*}$ & $-0.07^{*}$ & $-0.48^{*}$ & $-0.27^{*}$ & $-0.05^{*}$ & $-0.46^{*}$ & $+0.23^{*}$ & $-0.18^{*}$ & $-0.41^{*}$ \\
Tentativeness & $+0.01$ & $-0.21^{*}$ & $-0.03^{*}$ & $-0.19^{*}$ & $-0.21^{*}$ & $-0.04^{*}$ & $-0.20^{*}$ & $+0.18^{*}$ & $+0.01$ & $-0.16^{*}$ \\
Positive tone & $+0.01^{*}$ & $-0.00$ & $+0.08^{*}$ & $+0.18^{*}$ & $-0.01^{*}$ & $+0.07^{*}$ & $+0.17^{*}$ & $+0.08^{*}$ & $+0.18^{*}$ & $+0.10^{*}$ \\
\bottomrule
\end{tabular}}
\caption{Differences in selected linguistic effects, aggregated across responses from 600 questions, 12 languages, and eight models. Compares the five language pathways with each other: English (Eng), TTG, NL, OSP, and NP. }

\label{tab:app_llm_linguistic_overall_strategy}
\end{table*}

\subsection{Linguistic Effects by Language}
\label{app:llm_linguistic_by_language}

We also show in Table~\ref{tab:app_llm_linguistic_by_language} the breakdown of the four strategies compared to the English baseline by language group. 

\begin{table*}[t]
\centering\small
\begin{tabular}{llrrrr}
\toprule
Language group & Feature & TTG & NL & OSP & NP \\
\midrule
East Asian & Social attunement & $-0.11^{*}$ & $-0.46^{*}$ & $-0.16^{*}$ & $-0.55^{*}$ \\
 & Emotional expressiveness & $-0.07^{*}$ & $-0.40^{*}$ & $-0.13^{*}$ & $-0.48^{*}$ \\
 & Tentativeness & $+0.02$ & $-0.32^{*}$ & $-0.05^{*}$ & $-0.17^{*}$ \\
 & Positive tone & $+0.02$ & $-0.05^{*}$ & $+0.05^{*}$ & $+0.20^{*}$ \\
\midrule
Mid. East/Turkic & Social attunement & $-0.09^{*}$ & $-0.37^{*}$ & $-0.13^{*}$ & $-0.54^{*}$ \\
 & Emotional expressiveness & $-0.07^{*}$ & $-0.32^{*}$ & $-0.10^{*}$ & $-0.45^{*}$ \\
 & Tentativeness & $+0.02$ & $-0.26^{*}$ & $-0.03$ & $-0.21^{*}$ \\
 & Positive tone & $+0.04^{*}$ & $+0.08^{*}$ & $+0.15^{*}$ & $+0.21^{*}$ \\
\midrule
Germanic & Social attunement & $-0.06^{*}$ & $-0.37^{*}$ & $-0.08^{*}$ & $-0.60^{*}$ \\
 & Emotional expressiveness & $-0.00$ & $-0.27^{*}$ & $-0.03$ & $-0.49^{*}$ \\
 & Tentativeness & $+0.03$ & $-0.11^{*}$ & $-0.00$ & $-0.16^{*}$ \\
 & Positive tone & $+0.00$ & $+0.01$ & $+0.09^{*}$ & $+0.14^{*}$ \\
\midrule
Romance & Social attunement & $+0.00$ & $-0.31^{*}$ & $-0.10^{*}$ & $-0.66^{*}$ \\
 & Emotional expressiveness & $+0.04^{*}$ & $-0.20^{*}$ & $-0.02$ & $-0.51^{*}$ \\
 & Tentativeness & $-0.06^{*}$ & $-0.14^{*}$ & $-0.04^{*}$ & $-0.24^{*}$ \\
 & Positive tone & $-0.02$ & $-0.05^{*}$ & $+0.02$ & $+0.14^{*}$ \\
\bottomrule
\end{tabular}
\caption{Selected linguistic effects by language group and strategy, aggregated across responses from 600 questions and eight models.}
\label{tab:app_llm_linguistic_by_language}
\end{table*}

\subsection{Linguistic Effects by Topic}
\label{app:llm_linguistic_by_topic}
We include in Table~\ref{tab:app_llm_linguistic_by_topic} the breakdown of the four strategies compared to the English baseline by topic. 

\begin{table*}[t]
\centering\small
\begin{tabular}{llrrrr}
\toprule
Topic & Feature & TTG & NL & OSP & NP \\
\midrule
Work & Social attunement & $-0.39^{*}$ & $-0.64^{*}$ & $-0.45^{*}$ & $-0.98^{*}$ \\
 & Emotional expressiveness & $-0.53^{*}$ & $-0.68^{*}$ & $-0.57^{*}$ & $-1.10^{*}$ \\
 & Tentativeness & $-0.01$ & $-0.19^{*}$ & $-0.04^{*}$ & $-0.18^{*}$ \\
 & Positive tone & $-0.08^{*}$ & $-0.10^{*}$ & $-0.01$ & $+0.02$ \\
\midrule
Friends & Social attunement & $-0.00$ & $-0.32^{*}$ & $-0.04^{*}$ & $-0.62^{*}$ \\
 & Emotional expressiveness & $+0.09^{*}$ & $-0.24^{*}$ & $+0.02$ & $-0.46^{*}$ \\
 & Tentativeness & $+0.07^{*}$ & $-0.13^{*}$ & $+0.05^{*}$ & $-0.14^{*}$ \\
 & Positive tone & $+0.09^{*}$ & $+0.07^{*}$ & $+0.16^{*}$ & $+0.32^{*}$ \\
\midrule
Relationships & Social attunement & $+0.14^{*}$ & $-0.24^{*}$ & $+0.10^{*}$ & $-0.24^{*}$ \\
 & Emotional expressiveness & $+0.25^{*}$ & $-0.09^{*}$ & $+0.22^{*}$ & $-0.03$ \\
 & Tentativeness & $-0.04^{*}$ & $-0.24^{*}$ & $-0.07^{*}$ & $-0.25^{*}$ \\
 & Positive tone & $-0.01$ & $-0.00$ & $+0.06^{*}$ & $+0.16^{*}$ \\
\midrule
Family & Social attunement & $-0.00$ & $-0.32^{*}$ & $-0.07^{*}$ & $-0.51^{*}$ \\
 & Emotional expressiveness & $+0.10^{*}$ & $-0.18^{*}$ & $+0.04^{*}$ & $-0.34^{*}$ \\
 & Tentativeness & $-0.00$ & $-0.26^{*}$ & $-0.06^{*}$ & $-0.23^{*}$ \\
 & Positive tone & $+0.01$ & $+0.01$ & $+0.07^{*}$ & $+0.16^{*}$ \\
\bottomrule
\end{tabular}
\caption{Selected linguistic effects by topic and strategy, aggregated across responses from 12 languages and eight models.}
\label{tab:app_llm_linguistic_by_topic}
\end{table*}

\subsection{Linguistic Effects by LLM Model}
\label{app:llm_linguistic_by_llm}

Finally, we report model-level LLM-annotated linguistic scores relative to each model's English baseline in Table~\ref{tab:app_llm_linguistic_by_llm}. These results show that prompting-strategy effects vary substantially by model, suggesting that future work should examine how training data composition, model origin, alignment procedures, and multilingual coverage shape cross-cultural or multilingual behavior in LLMs.

\begin{table*}[t]
\centering\small
\begin{tabular}{llrrrr}
\toprule
Model & Feature & TTG & NL & OSP & NP \\
\midrule
Claude Opus-4.6 & Social attunement & $-0.02$ & $-0.14^{*}$ & $+0.05$ & $-0.10^{*}$ \\
 & Emotional expressiveness & $-0.02$ & $-0.16^{*}$ & $+0.04$ & $-0.10^{*}$ \\
 & Tentativeness & $+0.03$ & $-0.53^{*}$ & $-0.42^{*}$ & $-0.21^{*}$ \\
 & Positive tone & $+0.04$ & $+0.10^{*}$ & $+0.32^{*}$ & $+0.04$ \\
\midrule
GPT-4o & Social attunement & $-0.07^{*}$ & $-0.26^{*}$ & $-0.26^{*}$ & $-0.89^{*}$ \\
 & Emotional expressiveness & $-0.02$ & $-0.15^{*}$ & $-0.08^{*}$ & $-0.72^{*}$ \\
 & Tentativeness & $+0.06$ & $-0.15^{*}$ & $-0.09^{*}$ & $-0.50^{*}$ \\
 & Positive tone & $+0.04$ & $+0.02$ & $+0.01$ & $+0.28^{*}$ \\
\midrule
GPT-5.5 & Social attunement & $-0.21^{*}$ & $-0.71^{*}$ & $-0.02$ & $-0.26^{*}$ \\
 & Emotional expressiveness & $-0.18^{*}$ & $-0.42^{*}$ & $-0.04$ & $-0.17^{*}$ \\
 & Tentativeness & $-0.03$ & $-0.33^{*}$ & $-0.11^{*}$ & $-0.28^{*}$ \\
 & Positive tone & $+0.10^{*}$ & $-0.02$ & $+0.10^{*}$ & $+0.07$ \\
\midrule
Gemma-4-26B-A4B & Social attunement & $-0.08^{*}$ & $-0.44^{*}$ & $-0.15^{*}$ & $-0.64^{*}$ \\
 & Emotional expressiveness & $-0.05$ & $-0.37^{*}$ & $-0.13^{*}$ & $-0.55^{*}$ \\
 & Tentativeness & $+0.06^{*}$ & $+0.05^{*}$ & $+0.10^{*}$ & $-0.06^{*}$ \\
 & Positive tone & $+0.01$ & $-0.00$ & $+0.10^{*}$ & $+0.19^{*}$ \\
\midrule
Gemma-4-31B & Social attunement & $-0.08^{*}$ & $-0.38^{*}$ & $-0.14^{*}$ & $-0.88^{*}$ \\
 & Emotional expressiveness & $-0.02$ & $-0.36^{*}$ & $-0.10^{*}$ & $-0.77^{*}$ \\
 & Tentativeness & $+0.07^{*}$ & $+0.07^{*}$ & $+0.10^{*}$ & $-0.27^{*}$ \\
 & Positive tone & $+0.01$ & $+0.01$ & $+0.12^{*}$ & $+0.32^{*}$ \\
\midrule
Gemma-4-E4B & Social attunement & $+0.04$ & $-0.40^{*}$ & $-0.22^{*}$ & $-0.66^{*}$ \\
 & Emotional expressiveness & $+0.10^{*}$ & $-0.36^{*}$ & $-0.18^{*}$ & $-0.58^{*}$ \\
 & Tentativeness & $-0.11^{*}$ & $-0.19^{*}$ & $+0.22^{*}$ & $+0.21^{*}$ \\
 & Positive tone & $-0.06^{*}$ & $-0.02$ & $-0.02$ & $+0.26^{*}$ \\
\midrule
Qwen-3.6-27B & Social attunement & $-0.11^{*}$ & $-0.80^{*}$ & $-0.22^{*}$ & $-1.15^{*}$ \\
 & Emotional expressiveness & $-0.03$ & $-0.51^{*}$ & $-0.21^{*}$ & $-0.71^{*}$ \\
 & Tentativeness & $+0.01$ & $-0.16^{*}$ & $+0.01$ & $-0.21^{*}$ \\
 & Positive Tone & $-0.07^{*}$ & $-0.22^{*}$ & $-0.10^{*}$ & $-0.02$ \\
\midrule
Qwen-3.6-35B-A3B & Social attunement & $-0.03$ & $0.09^{*}$ & $+0.00$ & $-0.14^{*}$ \\
 & Emotional expressiveness & $-0.02$ & $-0.07^{*}$ & $+0.14^{*}$ & $-0.24^{*}$ \\
 & Tentativeness & $-0.03$ & $-0.41^{*}$ & $-0.05$ & $-0.22{*}$ \\
 & Positive Tone & $+0.01$ & $+0.13^{*}$ & $+0.19^{*}$ & $+0.30^{*}$ \\
\midrule
\bottomrule
\end{tabular}
\caption{Selected linguistic effects by model and strategy, aggregated across responses from 600 questions and 12 languages.}
\label{tab:app_llm_linguistic_by_llm}
\end{table*}

\subsection{Cross-Model Consistency Between NP and NL}
\label{app:cross_model}
\begin{figure}
    \centering
    \includegraphics[width=0.8\linewidth]{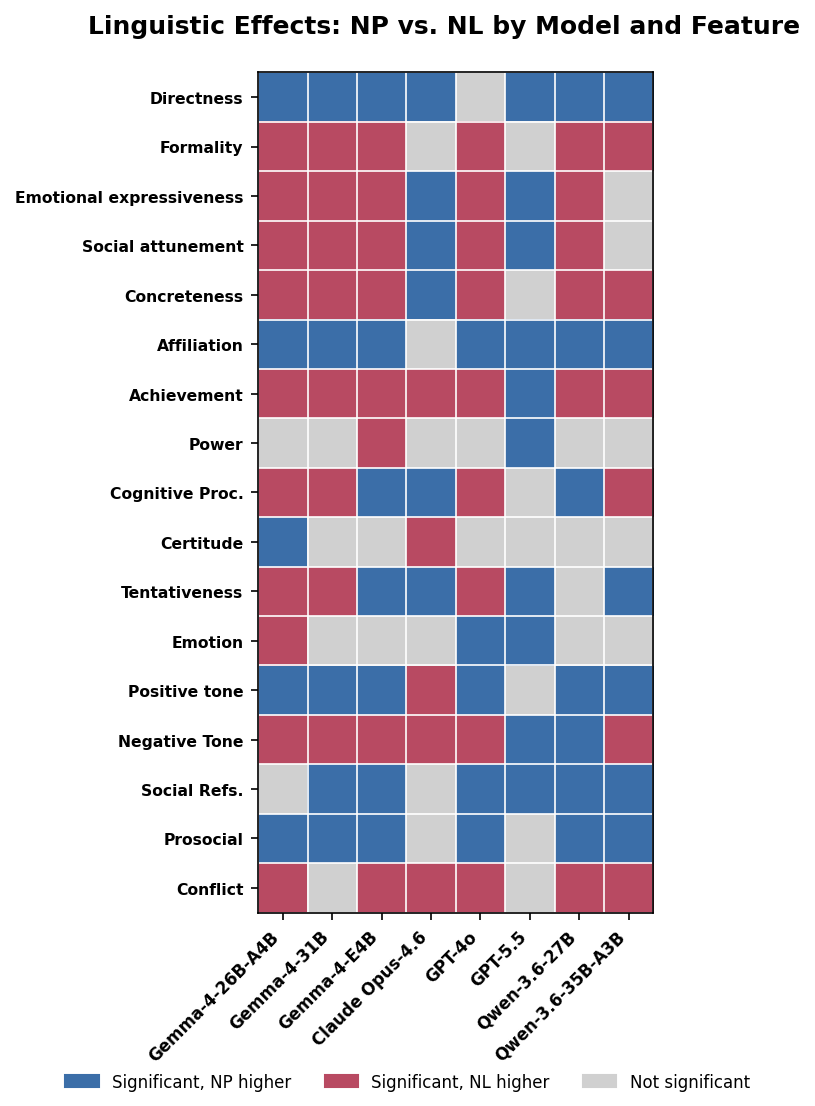}
    \caption{NP--NL direction of linguistic effects by model, aggregated across responses from 600 questions and 12 languages. Overall, the LLMs tend to be consistent in direction across models.
    }    
    \label{app:linguistic_consistency}
\end{figure}

We also include the cross-model differences between NP and NL in Figure~\ref{app:linguistic_consistency} to show that the language models have consistent patterns across linguistic features.

\subsection{Linguistic Effects by Language Group}
\label{app:linguistic_lang_group}

\begin{figure*}[t]
    \centering
    \includegraphics[width=\textwidth]{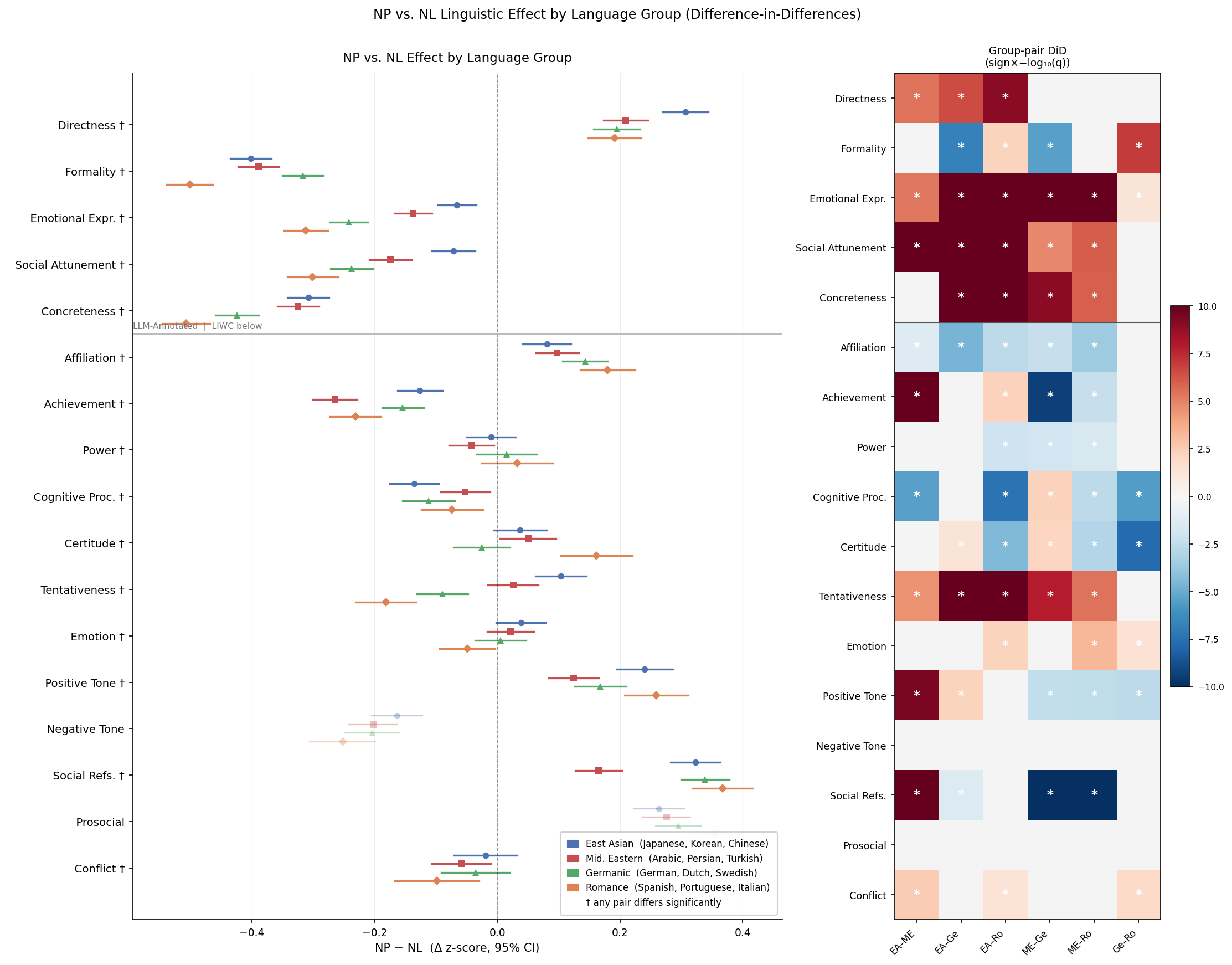}
    \caption{NP--NL differences in linguistic effects by language group, aggregated across responses from 600 questions and eight models. The left panel shows mean NP-minus-NL differences for LLM-annotated pragmatic features and LIWC lexical features, with 95\% confidence intervals. The right panel reports exploratory difference-in-differences tests comparing the NP--NL gap across language groups. 
    }
    \label{fig:app_linguistic_group_did}
\end{figure*}

Figure~\ref{fig:app_linguistic_group_did} reports NP--NL linguistic differences by broad language group. The main pattern is that the NP--NL difference is not uniform across language groups. Persona prompting does not introduce a single stable transformation of native-language generation. Instead, the size and direction of the gap depend on both the target-language group and the linguistic feature being measured.

\FloatBarrier

\section{Overall BCT Effects}
\label{app:bct_overall_strategy}

Table~\ref{tab:app_bct_overall_strategy} reports all pairwise differences in composite BCT across strategies. NP shows the largest overall departure from the English baseline, consistent with the linguistic results.

\begin{table*}[t]
\centering\small
\resizebox{\textwidth}{!}{%
\begin{tabular}{lrrrrrrrrrr}
\toprule
BCT composite & TTG--Eng & NL--Eng & OSP--Eng & NP--Eng & NL--TTG & OSP--TTG & NP--TTG & OSP--NL & NP--NL & NP--OSP \\
\midrule
BCT composite & $-0.09^{*}$ & $-0.54^{*}$ & $-0.32^{*}$ & $-0.98^{*}$ & $-0.45^{*}$ & $-0.23^{*}$ & $-0.88^{*}$ & $+0.22^{*}$ & $-0.44^{*}$ & $-0.65^{*}$ \\
\bottomrule
\end{tabular}}
\caption{Differences in overall composite BCT effects, aggregated across responses from 600 questions, 12 languages, eight models, and all BCT features. Compares the five language pathways with each other: English (Eng), TTG, NL, OSP, and NP. }
\label{tab:app_bct_overall_strategy}
\end{table*}

\subsection{BCT Effects by Language}
\label{app:bct_by_language}

Table~\ref{tab:app_bct_by_language} shows composite BCT strategy effects relative to the English baseline, pooled within each language group.

\begin{table*}[t]
\centering\small
\begin{tabular}{lrrrr}
\toprule
Language group & TTG & NL & OSP & NP \\
\midrule
East Asian & $-0.13^{*}$ & $-0.58^{*}$ & $-0.42^{*}$ & $-0.98^{*}$ \\
\midrule
Mid. East/Turkic & $-0.13^{*}$ & $-0.54^{*}$ & $-0.37^{*}$ & $-0.93^{*}$ \\
\midrule
Germanic & $-0.09^{*}$ & $-0.54^{*}$ & $-0.23^{*}$ & $-0.93^{*}$ \\
\midrule
Romance & $+0.00$ & $-0.51^{*}$ & $-0.25^{*}$ & $-1.08^{*}$ \\
\bottomrule
\end{tabular}
\caption{Overall composite BCT effects by language group and strategy, aggregated across responses from 600 questions, eight models, and all BCT features.}
\label{tab:app_bct_by_language}
\end{table*}

\subsection{BCT Effects by Topic}
\label{app:bct_by_topic}

Table~\ref{tab:app_bct_by_topic} presents composite BCT effects broken down by conversation topic.

\begin{table*}[t]
\centering\small
\begin{tabular}{lrrrr}
\toprule
Topic & TTG & NL & OSP & NP \\
\midrule
Work & $-0.18^{*}$ & $-0.61^{*}$ & $-0.40^{*}$ & $-1.13^{*}$ \\
\midrule
Friends & $-0.15^{*}$ & $-0.58^{*}$ & $-0.40^{*}$ & $-1.19^{*}$ \\
\midrule
Relationships & $+0.01$ & $-0.47^{*}$ & $-0.19^{*}$ & $-0.72^{*}$ \\
\midrule
Family & $-0.04^{*}$ & $-0.50^{*}$ & $-0.29^{*}$ & $-0.87^{*}$ \\
\bottomrule
\end{tabular}
\caption{Overall composite BCT effects by topic and strategy, aggregated across responses from 12 languages, eight models, and all BCT features.}
\label{tab:app_bct_by_topic}
\end{table*}

\subsection{BCT Effects by LLM Model}
\label{app:bct_by_llm}

Table~\ref{tab:app_bct_by_llm} reports model-level composite BCT effects relative to each model's own English baseline.

\begin{table*}[t]
\centering\small
\begin{tabular}{lrrrr}
\toprule
Model & TTG & NL & OSP & NP \\
\midrule
Claude Opus-4.6 & $-0.09^{*}$ & $-0.23^{*}$ & $-0.09^{*}$ & $-0.15^{*}$ \\
\midrule
GPT-4o & $-0.09^{*}$ & $-0.34^{*}$ & $-0.44^{*}$ & $-1.30^{*}$ \\
\midrule
GPT-5.5 & $-0.27^{*}$ & $-0.46^{*}$ & $-0.16^{*}$ & $-0.51^{*}$ \\
\midrule
Gemma-4-26B-A4B & $-0.09^{*}$ & $-0.79^{*}$ & $-0.37^{*}$ & $-1.14^{*}$ \\
\midrule
Gemma-4-31B & $-0.09^{*}$ & $-0.75^{*}$ & $-0.38^{*}$ & $-1.39^{*}$ \\
\midrule
Gemma-4-E4B & $+0.00$ & $-0.69^{*}$ & $-0.59^{*}$ & $-1.29^{*}$ \\
\midrule
Qwen-3.6-27B & $-0.09^{*}$ & $-0.57^{*}$ & $-0.33^{*}$ & $-1.08^{*}$ \\
\midrule
Qwen-3.6-35B-A3B & $-0.03$ & $-0.49^{*}$ & $-0.23^{*}$ & $-0.96^{*}$ \\
\bottomrule
\end{tabular}
\caption{Overall composite BCT effects by model and strategy, aggregated across responses from 600 questions, 12 languages, and all BCT features.}
\label{tab:app_bct_by_llm}
\end{table*}

\FloatBarrier

\section{Overall Action Distribution}
\label{app:action_overall_strategy}

Table~\ref{tab:app_action_overall_strategy} reports the forced-choice action distribution (Confrontation, Redirection, Disengagement) across all strategies. The English baseline is shown for reference; NL shows the largest shift, moving responses away from Confrontation and toward Redirection relative to English. Again, significance is computed using pairwise comparisons to the English baseline.

\begin{table*}[t]
\centering\small
\begin{tabular}{lrrr}
\toprule
Strategy & Confrontation & Redirection & Disengagement \\
\midrule
English & $34.1$ & $52.0$ & $13.9$ \\
\midrule
TTG & $35.4$ & $50.3$ & $14.3$ \\
NL & $28.8^{*}$ & $57.6^{*}$ & $13.5^{*}$ \\
OSP & $30.1^{*}$ & $54.6^{*}$ & $15.3^{*}$ \\
NP & $32.3$ & $53.4$ & $14.3$ \\
\bottomrule
\end{tabular}
\caption{Forced choice action rates for the five language pathways, aggregated across responses from 600 questions, 12 languages, and eight models.}
\label{tab:app_action_overall_strategy}
\end{table*}

\subsection{Action Distribution by Language}
\label{app:action_by_language}

Table~\ref{tab:app_action_by_language} breaks down the forced-choice action distribution by language group and strategy. 

\begin{table*}[t]
\centering\small
\begin{tabular}{llrrrr}
\toprule
Language group & Action & TTG & NL & OSP & NP \\
\midrule
East Asian & Confrontation & $33.0$ & $23.0^{*}$ & $25.0^{*}$ & $28.0^{*}$ \\
 & Redirection & $51.6$ & $61.2^{*}$ & $57.5^{*}$ & $57.0^{*}$ \\
 & Disengagement & $15.4$ & $15.8^{*}$ & $17.5^{*}$ & $15.0^{*}$ \\
\midrule
Mid.~East/Turkic & Confrontation & $36.1$ & $27.5^{*}$ & $28.7^{*}$ & $31.5^{*}$ \\
 & Redirection & $50.2$ & $58.7^{*}$ & $56.2^{*}$ & $55.3^{*}$ \\
 & Disengagement & $13.7$ & $13.8^{*}$ & $15.1^{*}$ & $13.2^{*}$ \\
\midrule
Germanic & Confrontation & $37.2^{*}$ & $34.1^{*}$ & $35.1$ & $37.0^{*}$ \\
 & Redirection & $49.3^{*}$ & $54.0^{*}$ & $51.7$ & $49.6^{*}$ \\
 & Disengagement & $13.5^{*}$ & $11.9^{*}$ & $13.3$ & $13.5^{*}$ \\
\midrule
Romance & Confrontation & $35.3$ & $30.7^{*}$ & $32.2$ & $33.4$ \\
 & Redirection & $50.2$ & $56.7^{*}$ & $52.5$ & $50.8$ \\
 & Disengagement & $14.5$ & $12.7^{*}$ & $15.2$ & $15.8$ \\
\bottomrule
\end{tabular}
\caption{Forced choice action rates by language group and strategy, aggregated across responses from 600 questions and eight models. }
\label{tab:app_action_by_language}
\end{table*}

\subsection{Action Distribution by Topic}
\label{app:action_by_topic}

Table~\ref{tab:app_action_by_topic} reports action distributions by conversation topic. 

\begin{figure*}
    \centering
    \includegraphics[width=1\linewidth]{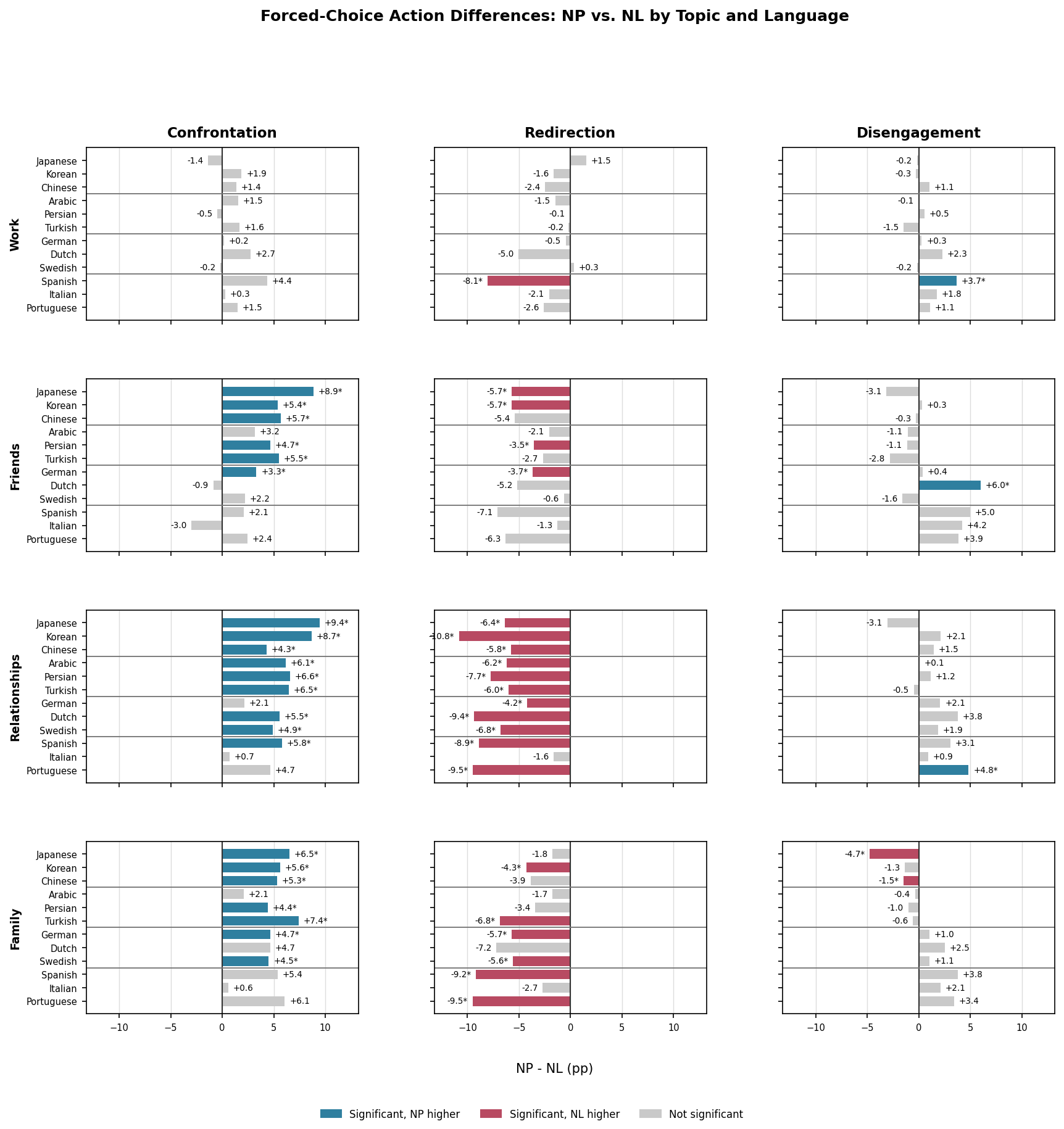}
    \caption{NP--NL differences in forced-choice action rates by topic and language, aggregated across responses from all eight models. With less data in each sample, many of the bars appear not significant. }
    \label{app:choice_by_topic}
\end{figure*}

We also report the NP--NL differences conditioned on topic and language in Figure~\ref{app:choice_by_topic}.

\begin{table*}[t]
\centering\small
\begin{tabular}{llrrrr}
\toprule
Topic & Action & TTG & NL & OSP & NP \\
\midrule
Work & Confrontation & $28.4$ & $23.8$ & $25.3$ & $24.7$ \\
 & Redirection & $57.8$ & $62.9$ & $60.2$ & $61.3$ \\
 & Disengagement & $13.8$ & $13.3$ & $14.5$ & $14.0$ \\
\midrule
Friends & Confrontation & $37.3$ & $32.2^{*}$ & $33.2^{*}$ & $35.4$ \\
 & Redirection & $44.8$ & $50.6^{*}$ & $47.3^{*}$ & $46.6$ \\
 & Disengagement & $17.9$ & $17.2^{*}$ & $19.6^{*}$ & $18.0$ \\
\midrule
Relationships & Confrontation & $44.0$ & $36.6^{*}$ & $37.6^{*}$ & $42.0$ \\
 & Redirection & $42.3$ & $50.6^{*}$ & $47.9^{*}$ & $43.6$ \\
 & Disengagement & $13.7$ & $12.9^{*}$ & $14.5^{*}$ & $14.5$ \\
\midrule
Family & Confrontation & $32.7$ & $23.6^{*}$ & $25.3^{*}$ & $28.1^{*}$ \\
 & Redirection & $55.5$ & $65.5^{*}$ & $61.7^{*}$ & $60.7^{*}$ \\
 & Disengagement & $11.8$ & $10.8^{*}$ & $13.0^{*}$ & $11.2^{*}$ \\
\bottomrule
\end{tabular}
\caption{Forced choice action rates by topic and strategy, aggregated across responses from 12 languages and eight models. }
\label{tab:app_action_by_topic}
\end{table*}

\subsection{Action Distribution by LLM Model}
\label{app:action_by_llm}

Table~\ref{tab:app_action_by_llm} reports model-level action distributions. 

\begin{table*}[t]
\centering\small
\begin{tabular}{llrrrr}
\toprule
Model & Action & TTG & NL & OSP & NP \\
\midrule
Claude Opus-4.6 & Confrontation & $38.4$ & $19.3^{*}$ & $24.3^{*}$ & $30.4$ \\
 & Redirection & $52.8$ & $74.3^{*}$ & $68.4^{*}$ & $62.2$ \\
 & Disengagement & $8.8$ & $6.4^{*}$ & $7.4^{*}$ & $7.4$ \\
\midrule
GPT-4o & Confrontation & $39.1$ & $31.7$ & $32.7$ & $37.1$ \\
 & Redirection & $45.9$ & $53.8$ & $53.1$ & $49.6$ \\
 & Disengagement & $15.0$ & $14.6$ & $14.3$ & $13.3$ \\
\midrule
GPT-5.5 & Confrontation & $35.2$ & $29.4$ & $31.1$ & $36.8$ \\
 & Redirection & $53.1$ & $57.8$ & $54.6$ & $50.9$ \\
 & Disengagement & $11.7$ & $12.8$ & $14.3$ & $12.3$ \\
\midrule
Gemma-4-26B-A4B & Confrontation & $26.1$ & $25.5$ & $24.5^{*}$ & $28.4$ \\
 & Redirection & $57.5$ & $56.7$ & $55.4^{*}$ & $55.6$ \\
 & Disengagement & $16.4$ & $17.9$ & $20.1^{*}$ & $16.0$ \\
\midrule
Gemma-4-31B & Confrontation & $48.9$ & $46.3$ & $45.9$ & $45.4$ \\
 & Redirection & $38.7$ & $41.7$ & $40.7$ & $42.1$ \\
 & Disengagement & $12.4$ & $12.0$ & $13.4$ & $12.5$ \\
\midrule
Gemma-4-E4B & Confrontation & $24.1^{*}$ & $18.7^{*}$ & $20.2$ & $18.0$ \\
 & Redirection & $57.9^{*}$ & $64.8^{*}$ & $60.8$ & $61.7$ \\
 & Disengagement & $18.0^{*}$ & $16.5^{*}$ & $19.0$ & $20.3$ \\
\midrule
Qwen-3.6-27B & Confrontation & $34.6$ & $29.0$ & $30.1$ & $30.3$ \\
 & Redirection & $49.5$ & $57.2$ & $53.0$ & $53.6$ \\
 & Disengagement & $15.9$ & $13.8$ & $16.9$ & $16.1$ \\
\midrule
Qwen-3.6-35B-A3B & Confrontation & $36.8$ & $30.8$ & $32.0$ & $32.0$ \\
 & Redirection & $47.0$ & $54.8$ & $50.8$ & $51.5$ \\
 & Disengagement & $16.2$ & $14.3$ & $17.2$ & $16.5$ \\
\bottomrule
\end{tabular}
\caption{Forced choice action rates by model and strategy, aggregated across responses from 600 questions and 12 languages. }
\label{tab:app_action_by_llm}
\end{table*}

\FloatBarrier

\section{Sensitivity to Persona-Prompt Wording}
\label{app:prompt_sensitivity}

We test whether the main NP--NL findings depend on the exact wording of the native-speaker persona. In addition to the original NP formulation, we evaluate two controlled variants that modify only the native-speaker identity clause:

\begin{itemize}
\item \textbf{Original NP:} "You are a native \{Language\} speaker."
    \item \textbf{Grew-up NP:} "You are a native \{Language\} speaker who grew up using that language."
\item \textbf{Current-use NP:} "You are a native \{Language\} speaker who currently uses that language."
\end{itemize}

We evaluate the three formulations on a stratified subset of 100 prompts, with 25 prompts per topic, across the 12 non-English languages and two models: \texttt{claude-opus-4.6} and \texttt{google/gemma-4-E4B-it}. We use the same matched comparisons and significance-testing procedures as in the main analysis. The full NP and NL generation procedures are described earlier in Appendix~\ref{app:generation_prompts}.

\begin{figure}
\centering
\includegraphics[width=0.5\textwidth]{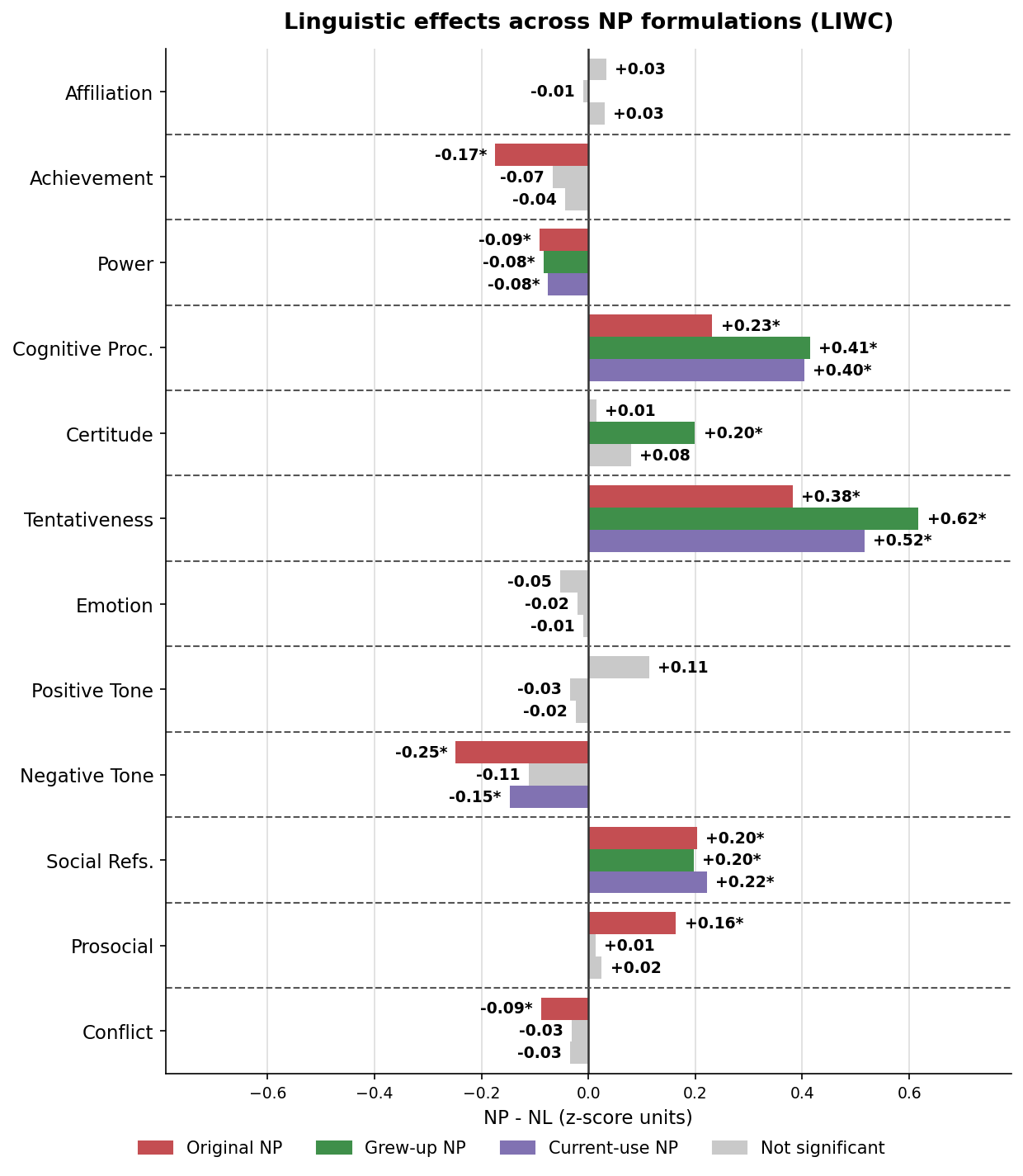}
\caption{NP--NL differences in LIWC linguistic features by persona formulations, aggregated across responses from 100 questions, 12 languages, and two models. }
\label{fig:app_prompt_sensitivity_liwc}
\end{figure}

Figures~\ref{fig:app_prompt_sensitivity_llm} and~\ref{fig:app_prompt_sensitivity_liwc} report the linguistic results. Several effects recur under all three formulations. Relative to NL, each NP formulation is significantly more direct, less formal, and less concrete. All three also use significantly less power-related language and more cognitive-process language, tentativeness, and social references. Other effects are more sensitive to wording: reductions in emotional expressiveness and social attunement weaken under Current-use NP, while effects for achievement, certitude, prosocial language, and conflict vary in significance across formulations. These results are consistent with the broader linguistic patterns.

\begin{figure}
\centering
\includegraphics[width=0.5\textwidth]{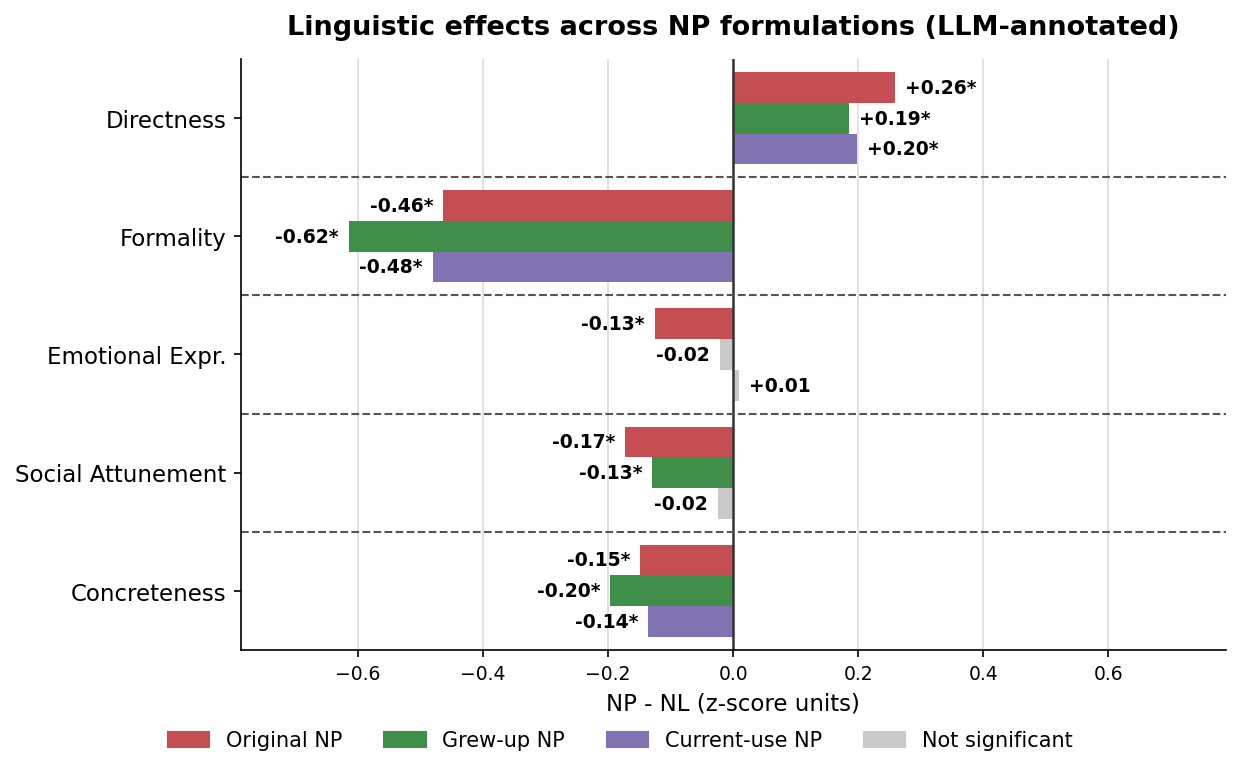}
\caption{NP--NL differences in LLM-annotated linguistic features by persona formulations, aggregated across responses from 100 questions, 12 languages, and two models. }
\label{fig:app_prompt_sensitivity_llm}
\end{figure}

\begin{figure}
\centering
\includegraphics[width=0.5\textwidth]{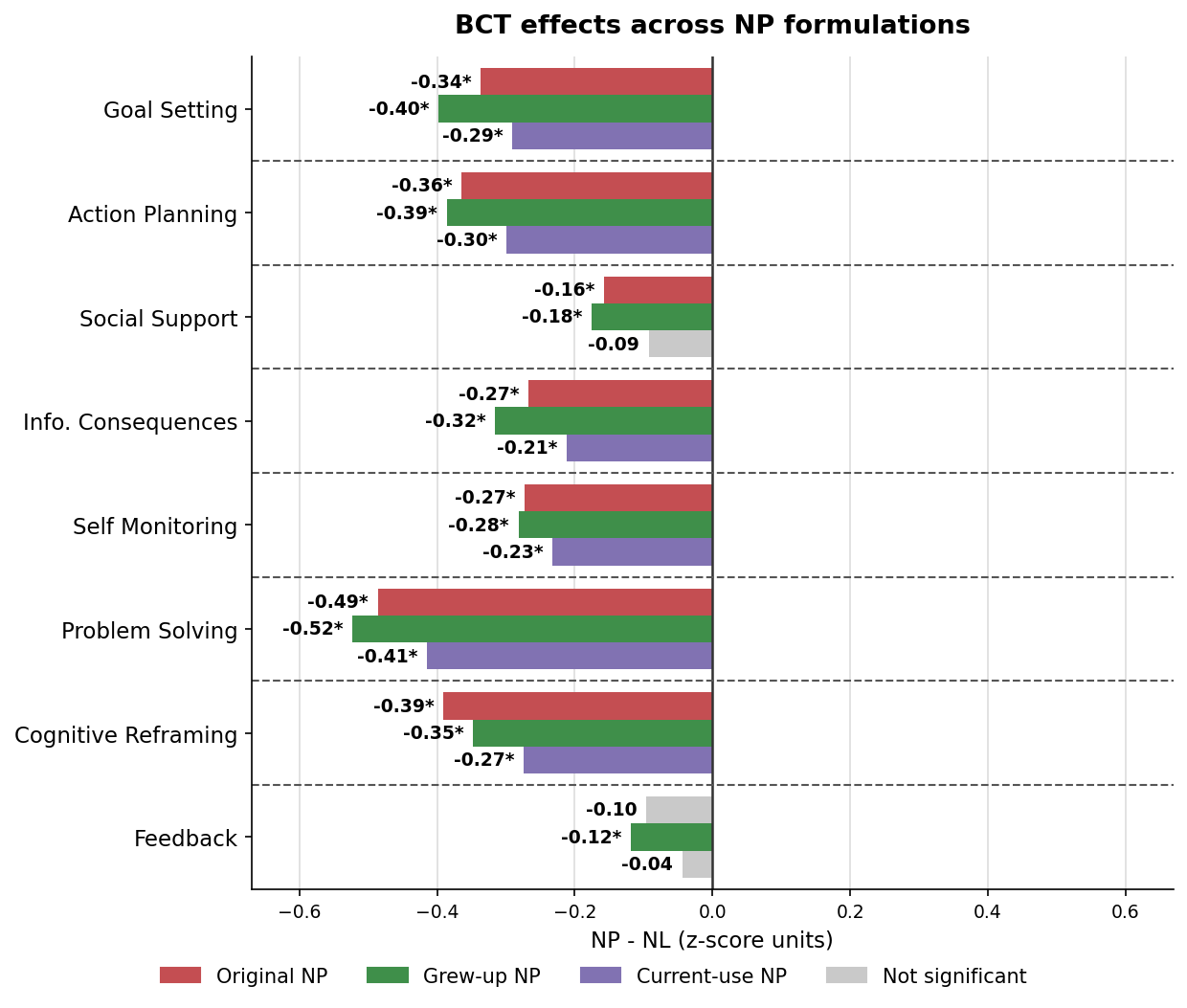}
\caption{NP--NL differences in BCT features by persona formulations, aggregated across responses from 100 questions, 12 languages, and two models. }
\label{fig:app_prompt_sensitivity_bct}
\end{figure}

Figure~\ref{fig:app_prompt_sensitivity_bct} shows that all eight BCT dimensions shift in the same direction under every formulation: each NP variant provides less behavioral scaffolding than NL. Six dimensions are significant under all three formulations, including goal setting, action planning, consequence reasoning, self-monitoring, problem solving, and cognitive reframing. 

\begin{figure}
    \centering
    \includegraphics[width=0.5\textwidth]{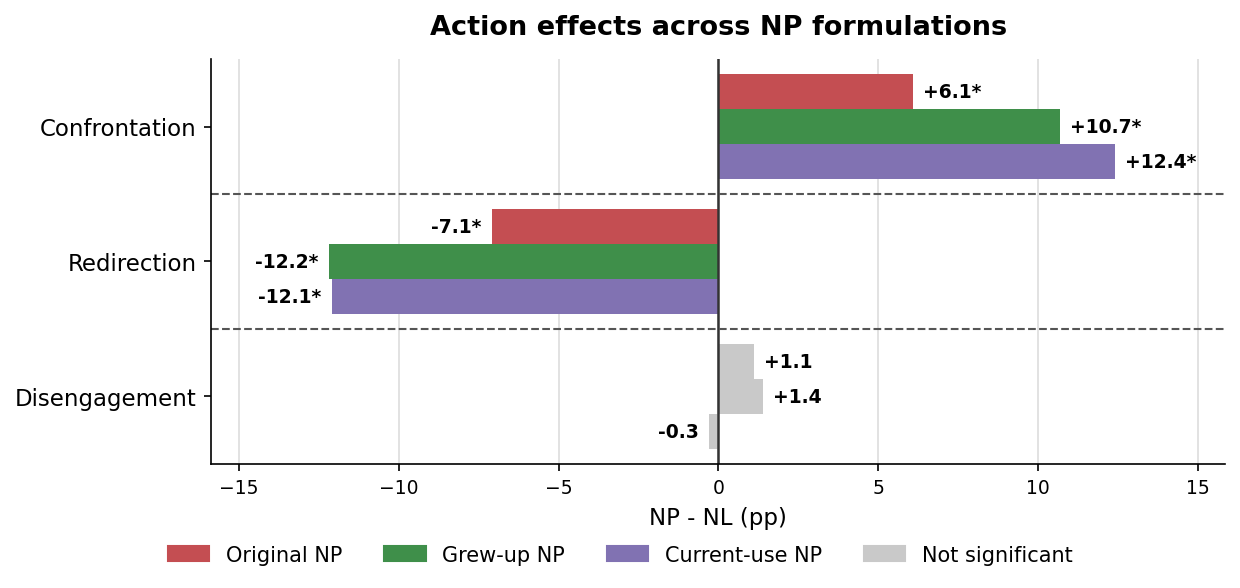}
    \caption{NP--NL differences in forced-choice action rates by persona formulations, aggregated across responses from 100 questions, 12 languages, and two models. }
    \label{fig:app_prompt_sensitivity}
\end{figure}

Forced-choice effects vary more in magnitude, but all three formulations remain more confrontation-oriented and less redirection-oriented than NL, as shown in Figure~\ref{fig:app_prompt_sensitivity}. Overall, the principal directional findings persist across the controlled wording changes, although their magnitudes and statistical significance remain prompt-sensitive.

\FloatBarrier

\section{Sensitivity to Translation Quality}
\label{app:translation_sensitivity}

We further assess whether the observed NP--NL differences could be driven by errors in the translated prompts. We repeat the main analyses after retaining only prompt--language pairs whose selected translation has a mean back-translation similarity of at least 0.95. The filter is applied at the prompt--language level before comparing NP and NL responses and retains 4,015 of 7,200 pairs (55.8 percent). Table~\ref{tab:app_translation_retention} reports the retained sample by language.

\begin{table*}[t]
\centering
\small
\begin{tabular}{lrr}
\toprule
Language & Retained pairs & Retained (\%) \\
\midrule
Japanese   & 139 & 23.2 \\
Korean     & 261 & 43.5 \\
Chinese    & 178 & 29.7 \\
Arabic     & 364 & 60.7 \\
Persian    & 273 & 45.5 \\
Turkish    & 306 & 51.0 \\
German     & 390 & 65.0 \\
Dutch      & 413 & 68.8 \\
Swedish    & 379 & 63.2 \\
Spanish    & 463 & 77.2 \\
Italian    & 414 & 69.0 \\
Portuguese & 435 & 72.5 \\
\midrule
Overall    & 4,015 & 55.8 \\
\bottomrule
\end{tabular}
\caption{Prompt--language pairs retained by the translation-quality filter. Each language has 600 pairs before filtering. Retention requires mean back-translation similarity $\geq 0.95$.}
\label{tab:app_translation_retention}
\end{table*}

Because retention rates differ across languages, directly pooling the filtered observations would change the language composition of the analysis. We therefore estimate NP--NL contrasts separately within each language, using the same matched prompt--model comparisons as in the main analysis. 

\begin{figure*}[t]
\centering
\includegraphics[width=\textwidth]{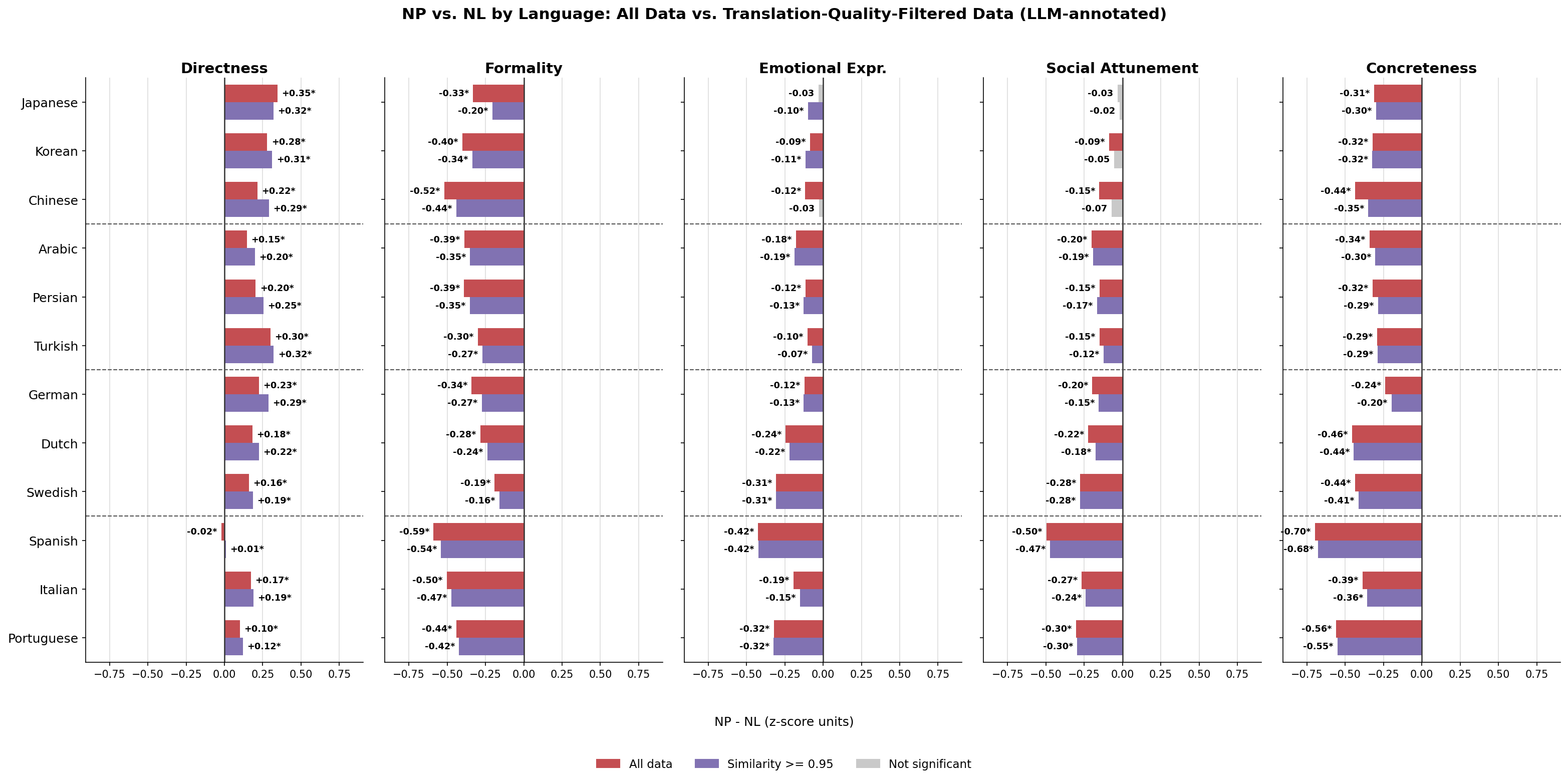}
\caption{NP--NL differences in LLM-annotated linguistic features by language and translation quality, aggregated across responses from 600 questions and eight models. }
\label{fig:app_translation_llm}
\end{figure*}

Figure~\ref{fig:app_translation_llm} compares the five LLM-annotated dimensions before and after filtering. Of the 60 language--dimension contrasts, 59 retain their direction; the mean absolute change is approximately 0.03 $z$-score units, and the largest change is 0.13. The broader pattern remains consistent: NP is generally more direct, while NL is generally more formal, emotionally expressive, socially attuned, and concrete.

\begin{figure*}[t]
\centering
\includegraphics[width=\textwidth]{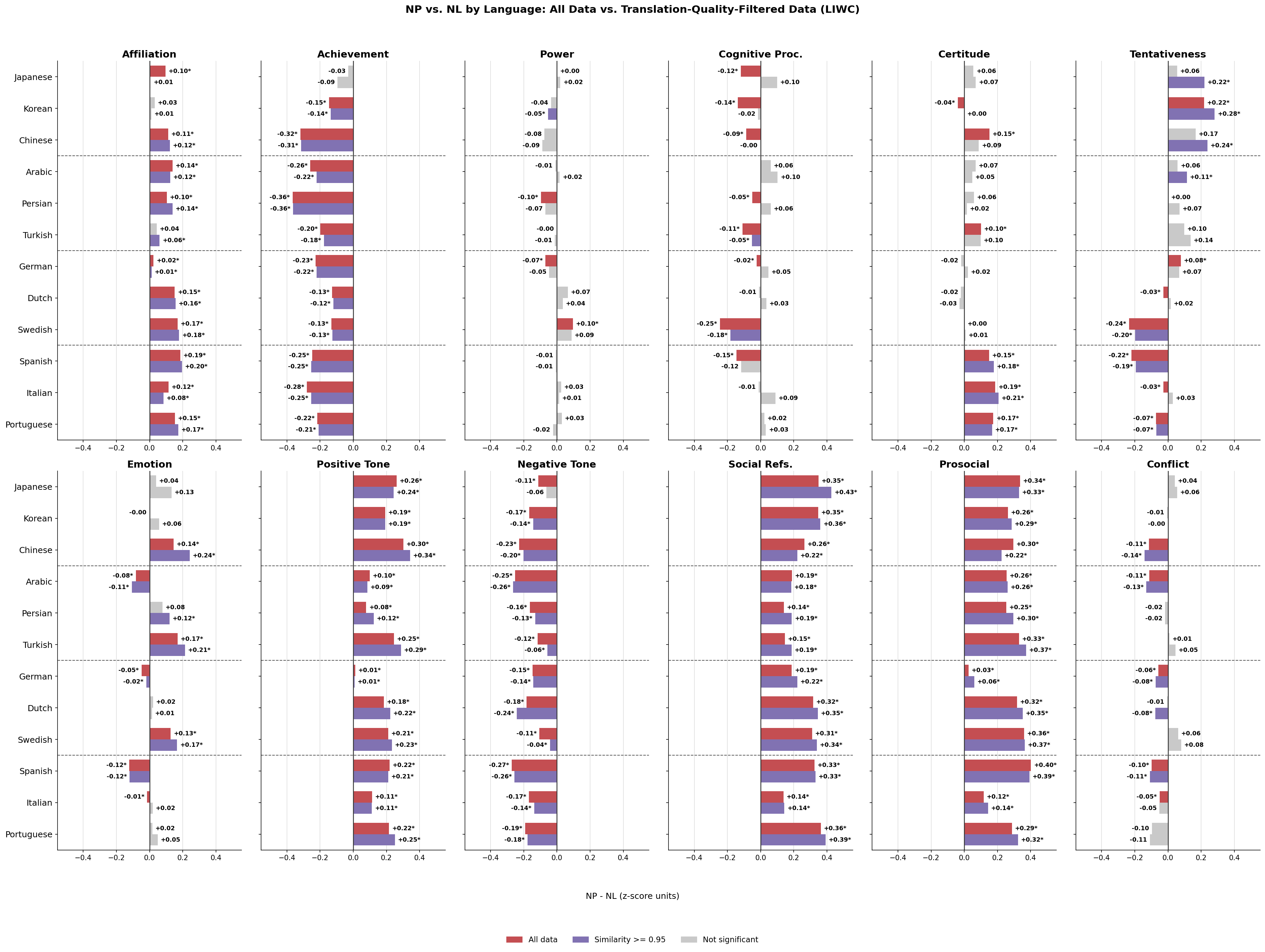}
\caption{NP--NL differences in LIWC linguistic features by language and translation quality, aggregated across responses from 600 questions and eight models. }
\label{fig:app_translation_liwc}
\end{figure*}

The LIWC results are also directionally stable (Figure~\ref{fig:app_translation_liwc}). Of the 144 language--dimension contrasts, 131 retain their direction. Twelve of the 13 reversals involve contrasts with estimates no larger than 0.09 in absolute value, indicating fluctuation around near-zero differences rather than substantive changes. The more pronounced lexical patterns remain: NP generally uses more affiliation, positive tone, social references, and prosocial language, while using less achievement and negative-tone language than NL. Most instability occurs among dimensions for which the original NP--NL differences are already small.

\begin{figure*}[t]
\centering
\includegraphics[width=\textwidth]{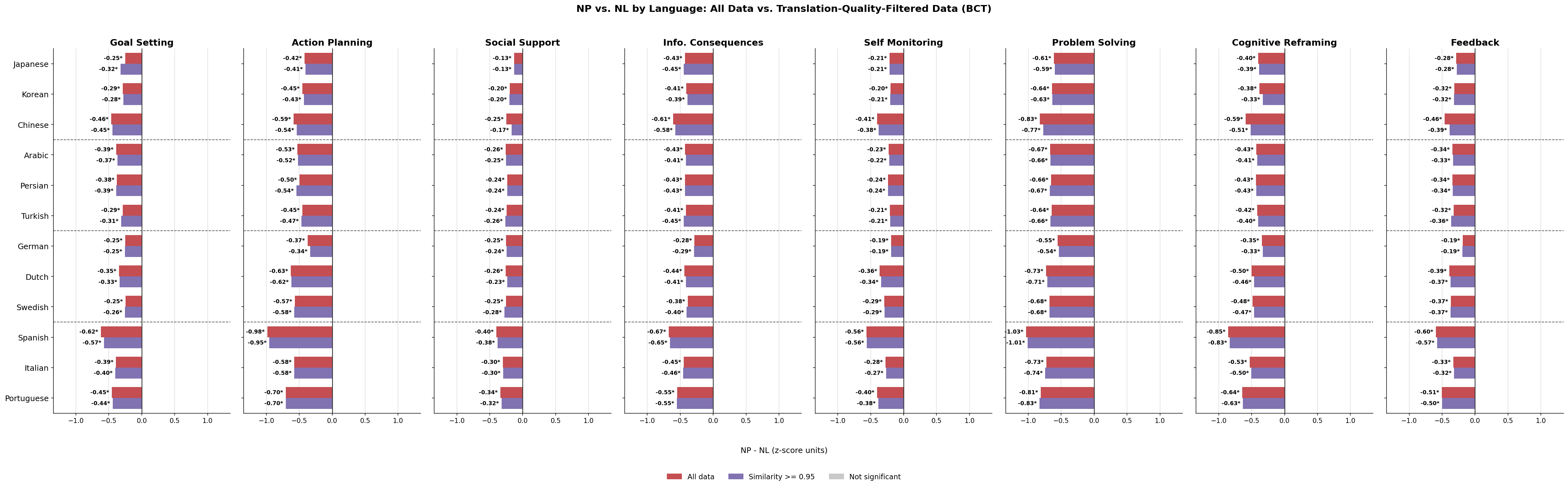}
\caption{NP--NL differences in BCT features by language and translation quality, aggregated across responses from 600 questions and eight models. }
\label{fig:app_translation_bct}
\end{figure*}

Figure~\ref{fig:app_translation_bct} reports the eight BCT dimensions. All 96 language--dimension contrasts remain negative after filtering, indicating less behavioral scaffolding under NP than under NL in every language and dimension. The mean absolute change is approximately 0.02 $z$-score units, with a maximum change of 0.09. 

\begin{figure*}[t]
\centering
\includegraphics[width=0.75\textwidth]{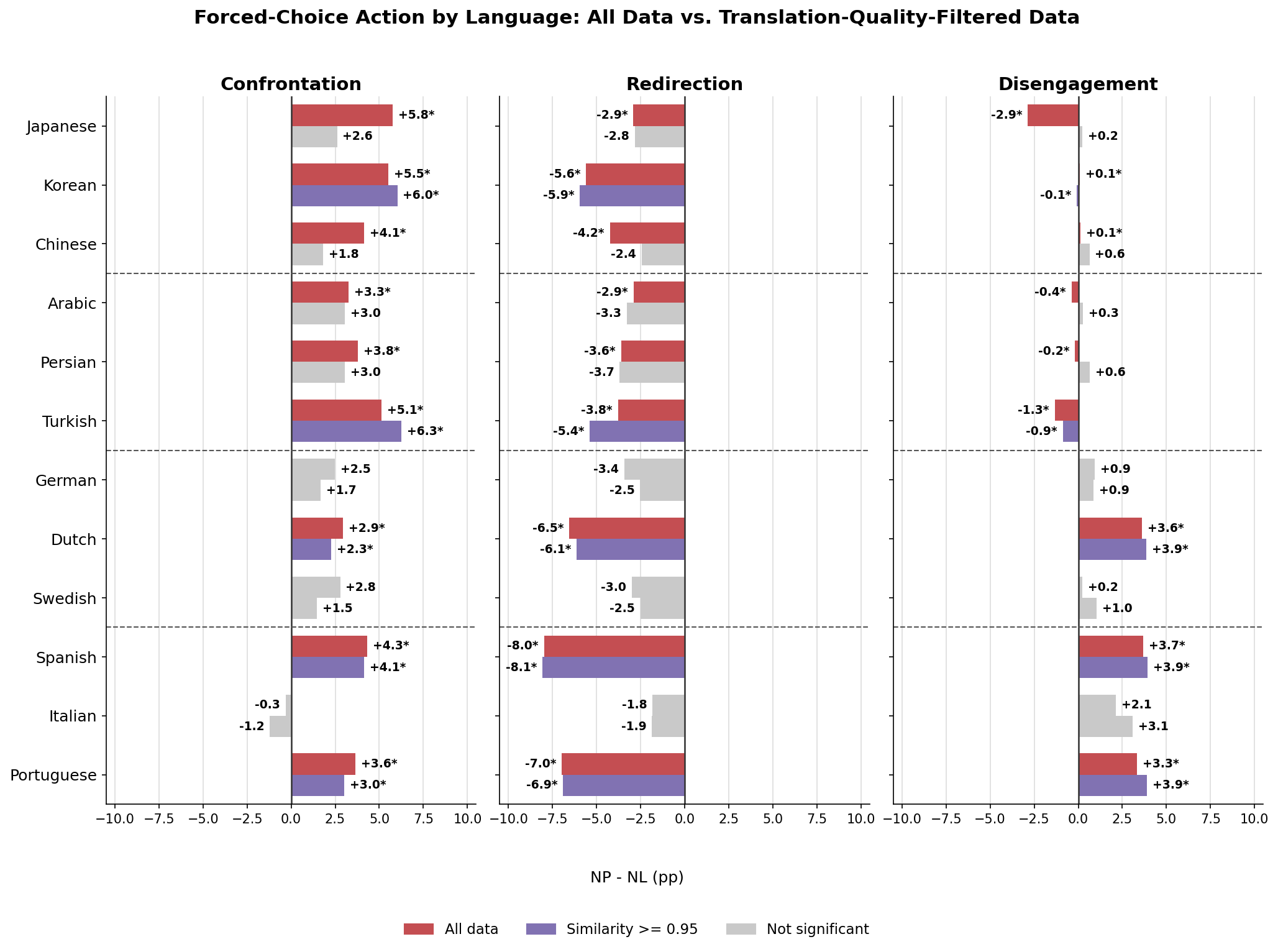}
\caption{NP--NL differences in forced-choice action rates features by language and translation quality, aggregated across responses from 600 questions and eight models. }
\label{fig:app_translation_action}
\end{figure*}

The forced-choice results preserve the main aggregate pattern (Figure~\ref{fig:app_translation_action}) as well. At the language level, the confrontation and redirection contrasts retain their directions in all 12 languages, whereas the smaller base disengagement contrasts are less stable.

Overall, restricting the analysis to high-similarity prompt translations produces little change in the principal NP--NL patterns. Nearly all linguistic contrasts retain their direction, every BCT contrast remains negative, and the shift from redirection toward confrontation persists.

\FloatBarrier

\section{Cross-LLM-as-a-Judge Comparison}
\label{app:cross_llm}
We test whether the LLM-annotated results depend on the choice of automated evaluator. In addition to the original \texttt{gpt-4o} annotations, we annotate the same responses using \texttt{claude-sonnet-5} with identical linguistic-style and BCT rubrics. We evaluate a stratified subset of 100 prompts, with 25 prompts per topic, across the 12 non-English languages and three generation models: \texttt{gpt-4o}, \texttt{claude-opus-4.6}, and \texttt{google/gemma-4-E4B-it}. We use the same matched comparisons and significance-testing procedure as in the main analysis. Because LIWC features and forced-choice outcomes do not rely on an LLM evaluator, this robustness check applies only to the five linguistic-style and eight BCT dimensions.

\begin{figure}
\centering
\includegraphics[width=0.5\textwidth]{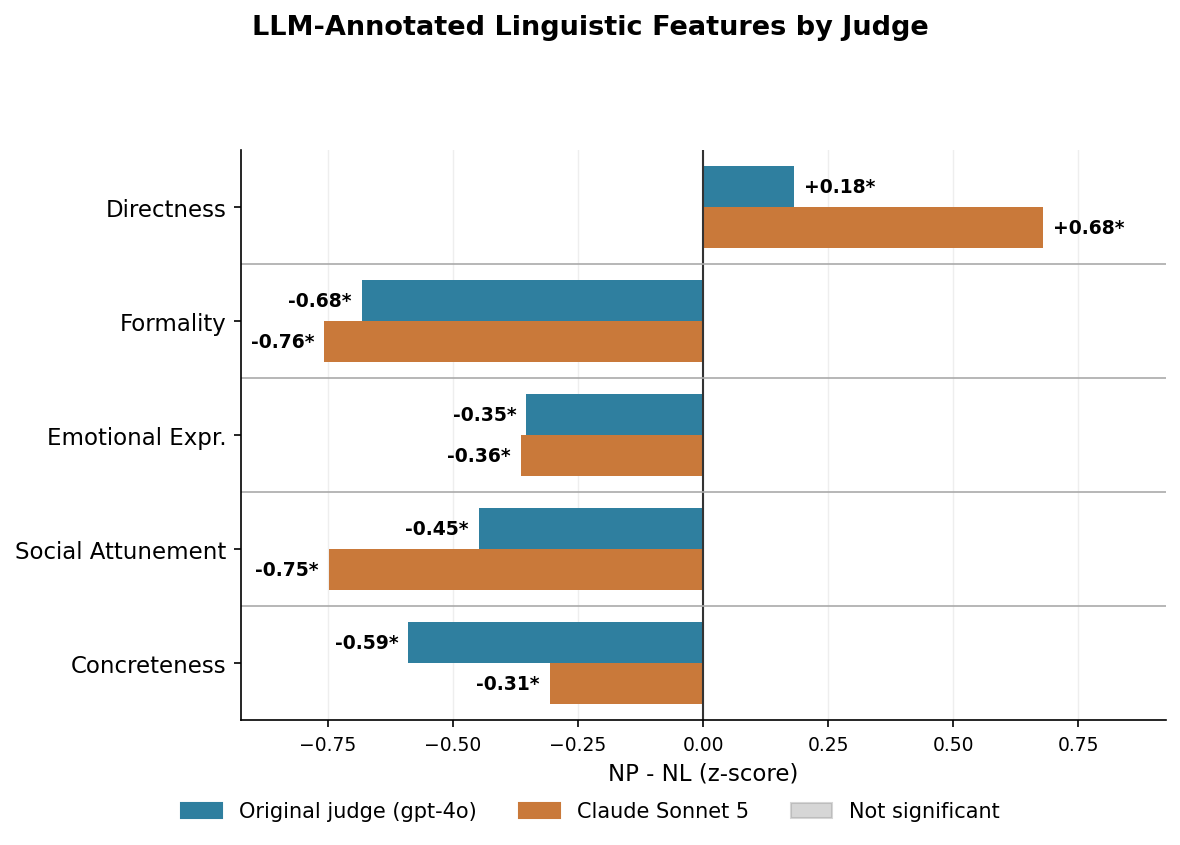}
\caption{NP--NL differences in LLM-annotated linguistic features by LLM-as-a-judge model, aggregated across responses from 100 questions, 12 languages, and three models. }
\label{fig:cross_llm_linguistic}
\end{figure}

Figure~\ref{fig:cross_llm_linguistic} compares the linguistic-style effects estimated by the two judges. At the aggregate level, the judges agree on the direction of all five NP--NL effects. At the language level, they agree on 58 of the 60 language-by-dimension effects ($96.7\%$), and their standardized effect sizes are strongly correlated ($r=.89$). When separated by generation model, the judges agree on 13 of the 15 model-by-dimension effects, with similarly correlated effect sizes ($r=.90$). Thus, the main linguistic differences are largely stable across evaluators.

\begin{figure}
\centering
\includegraphics[width=0.5\textwidth]{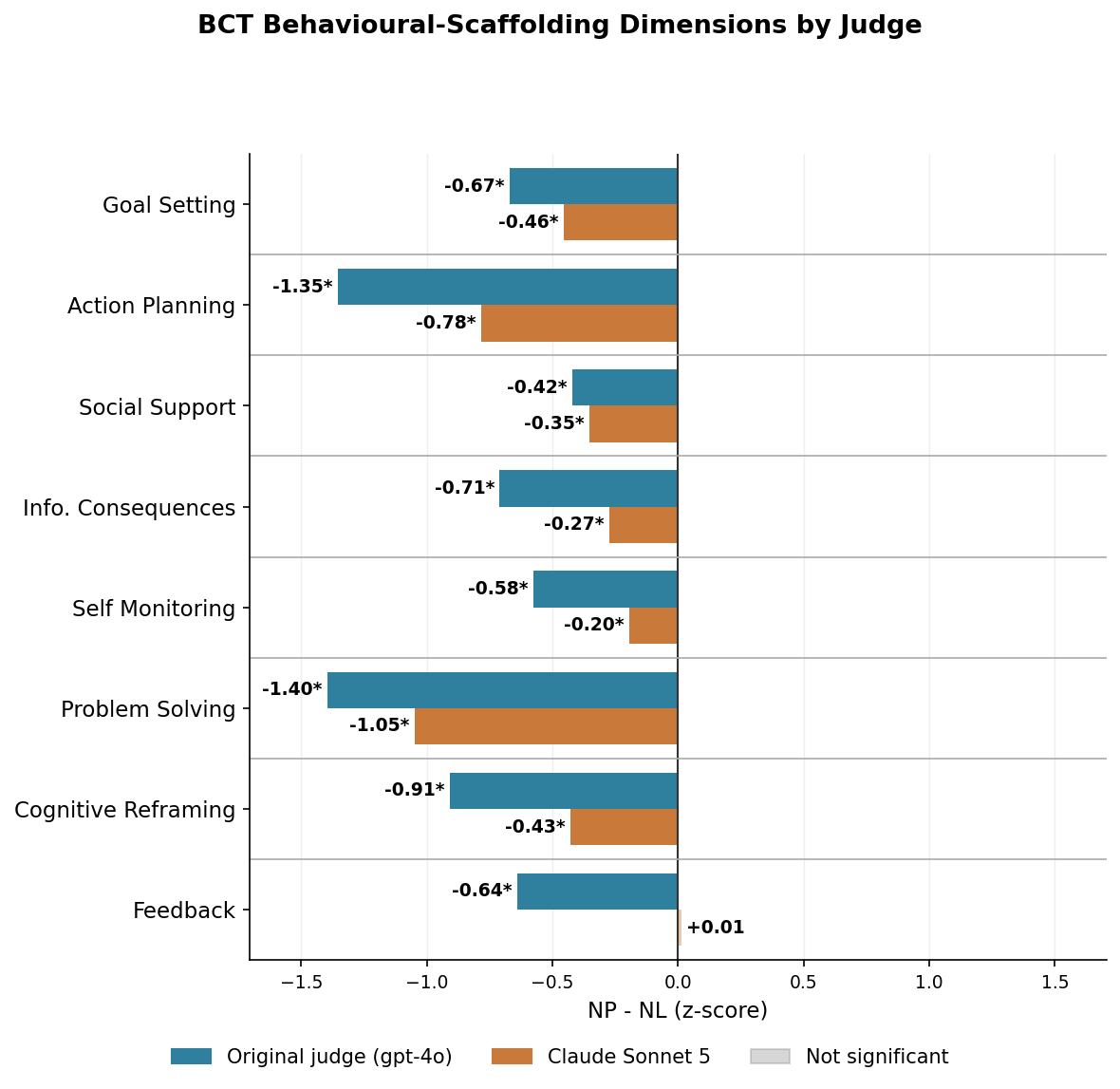}
\caption{NP--NL differences in BCT features by LLM-as-a-judge model, aggregated across responses from 100 questions, 12 languages, and three models. }
\label{fig:cross_llm_bct}
\end{figure}

Figure~\ref{fig:cross_llm_bct} reports the BCT results. The judges agree on the direction of seven of the eight aggregate NP--NL effects. These seven dimensions indicate less behavioral scaffolding under NP than under NL and are significant under both evaluators. The only discrepancy is feedback: \texttt{gpt-4o} estimates a moderate NP--NL reduction, whereas \texttt{claude-sonnet-5} estimates a near-null effect.

\begin{figure}
\centering
\includegraphics[width=0.5\textwidth]{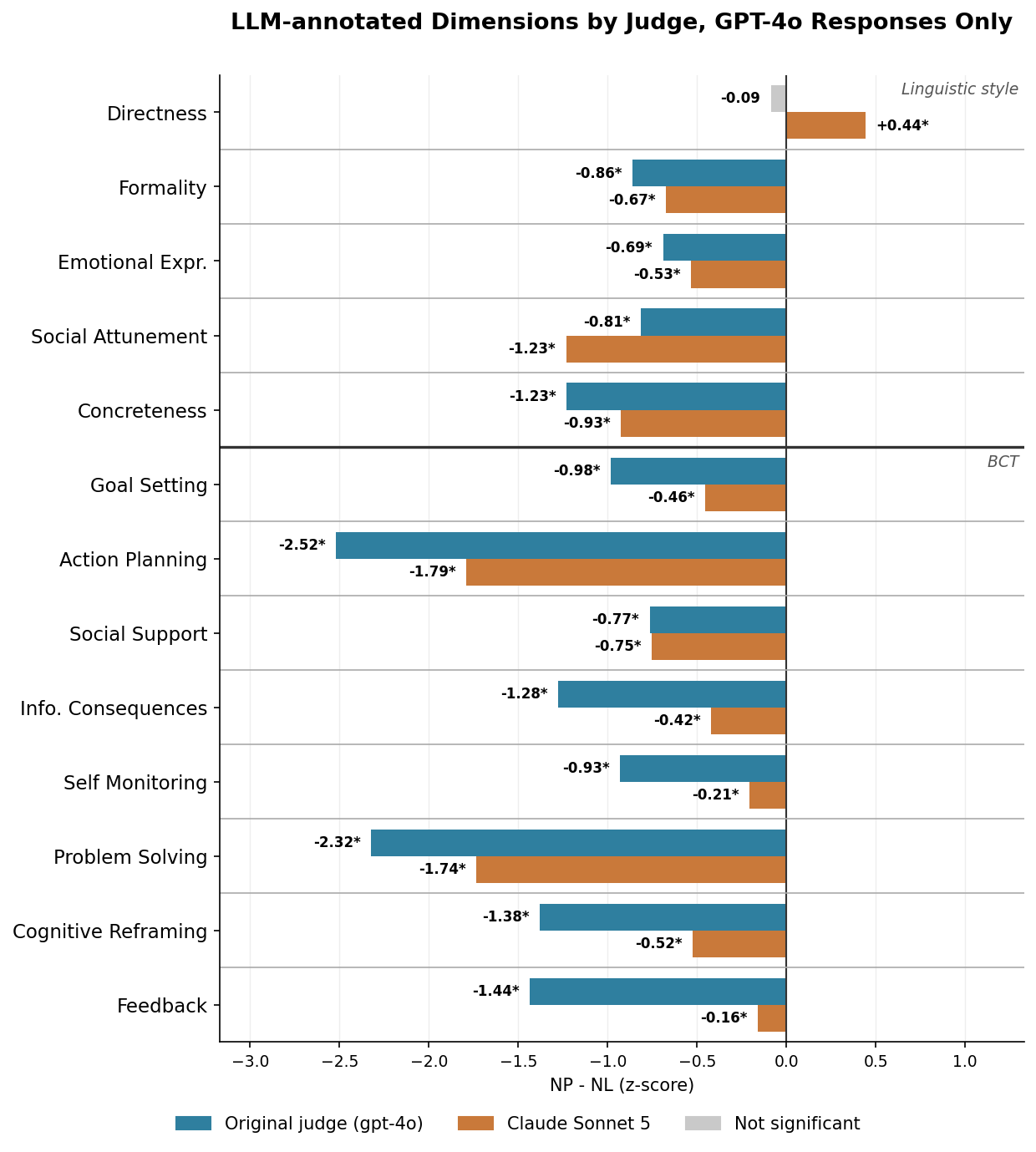}
\caption{NP--NL differences in LLM-annotated (linguistic and BCT) features on \texttt{gpt-4o} generated responses, aggregated across responses from 100 questions and 12 languages. }\label{fig:cross_llm_gpt}
\end{figure}

Agreement is also strong when results are separated by generation model. Critically, agreement does not weaken for responses generated by \texttt{gpt-4o}, despite \texttt{gpt-4o} serving as the original evaluator (Figure~\ref{fig:cross_llm_gpt}). For these responses, the judges agree on 12 of the 13 effects ($r=.77$). The only directional difference is directness, for which the original evaluator estimates a near-null effect while \texttt{claude-sonnet-5} estimates a positive effect.

Overall, the principal directional findings persist across the two automated evaluators, including when \texttt{gpt-4o} evaluates its own generated responses.

\end{document}